\documentclass[preprint,12pt]{elsarticle}

\usepackage{amssymb}
\usepackage{natbib}
\usepackage{times}
\usepackage{helvet}
\usepackage{courier}
\usepackage{float} 
\usepackage{array}
\usepackage{longtable} 
\usepackage{amsmath}
\usepackage{graphicx}
\usepackage{epstopdf}
\usepackage{booktabs}
\usepackage{makecell}
\usepackage{multirow}
\usepackage{graphicx,subfigure}
\usepackage{diagbox}
\usepackage{rotating}
\usepackage{geometry}
\usepackage{caption}
\usepackage{color}
\newtheorem{myDef}{Definition}

\usepackage{multirow}

\usepackage[table,xcdraw]{xcolor}

\usepackage{algorithmic}
\usepackage[linesnumbered,ruled]{algorithm2e}

\usepackage{threeparttable} 
\usepackage[colorlinks,linkcolor=blue,anchorcolor=red,citecolor=blue]{hyperref}
\graphicspath{{fig/}}

\journal{Information Sciences}

\begin{document}
\begin{frontmatter}



\title{Investigation of Bare-bones Algorithms from Quantum Perspective: A Quantum Dynamical
Global Optimizer}

\author[focal1]{Peng Wang}
\author[focal1]{Gang Xin\corref{cor1}}
\ead{xin\_gang@swun.edu.cn}
\author[focal1]{Fang Wang}


\cortext[cor1]{Corresponding author}

\address[focal1]{School of Computer Science and Technology, Southwest Minzu University, Chengdu 610225, China}

\begin{abstract}

Recent decades, the emergence of numerous novel algorithms makes it a gimmick to propose an intelligent optimization system based on metaphor, and hinders researchers from exploring the essence of search behavior in algorithms. However, it is difficult to directly discuss the search behavior of an intelligent optimization algorithm, since there are so many kinds of intelligent schemes. To address this problem, an intelligent optimization system is regarded as a simulated physical optimization system in this paper. The dynamic search behavior of such a simplified physical optimization system are investigated with quantum theory. To achieve this goal, the Schr\"odinger equation is employed as the dynamics equation of the optimization algorithm, which is used to describe dynamic search behaviours in the evolution process with quantum theory. Moreover, to explore the basic behaviour of the optimization system, the optimization problem is assumed to be decomposed and approximated. Correspondingly, the basic search behaviour is derived, which constitutes the basic iterative process of a simple optimization system. The basic iterative process is compared with some classical bare-bones schemes to verify the similarity of search behavior under different metaphors. The search strategies of these bare bones algorithms are analyzed through experiments.

\end{abstract}


\end{frontmatter}

\section{Introduction}

In 1975, Professor Holland of the University of Michigan formally put forward a genetic algorithm in his book \textit{Adaptation of natural and artificial systems} and proved the schema theorem of the genetic algorithm (GA) \cite{Holl1975}. The scheme opened the door for intelligent optimization systems and the concepts of natural systems are employed to explain the optimization process of algorithms, which greatly promoted the development of algorithms and led to the growth of this field. Due to the wide range of application scenarios of intelligent algorithms, this field has become the focus of numerous scholars, and the emergence of new methods has gradually become endless and dazzling \cite{fister2016new}\cite{weyland2010rigorous}. In those novel algorithms, the interpretation of the evolutionary process is sometimes complicated and unconvincing. Since the basic characteristic of an intelligent optimization system is the evolutionary processes that evolve over time, in which its motion state is the result of the interaction of search strategies and sampled results of the objective function, the research on the basic dynamic mechanism of the algorithm has attracted the attention of many researchers, and it has become a vital issue in developing intelligent algorithms. However, this is not a simple work, due to the variety of algorithm models. Fortunately, there is still some work on the bare-bones model of classical algorithm  that can be used for reference.

Kennedy, the inventor of particle swarm optimization (PSO) \cite{kennedy1995particle}, who realized that optimization algorithms have a simple model and basic iterative operations in common, proposed a corresponding "bare-bones" edition of PSO. He hoped to use this basic algorithm to reveal the mystery behind the similarity of swarm optimization algorithms and propose new methods \cite{kennedy2003bare}\cite{kennedy2004probability}. Similarly, there are many such explorations. Omran proposed the bare-bones differential evolution (BBDE) algorithm to simplify the parameters of the differential evolution (DE) algorithm \cite{storn1997differential} \cite{omran2009bare}. Later, Wang and Omran further simplified the structure and proposed Gaussian bare-bone differential evolution (GBDE) \cite{wang2013gaussian}. Tan, the inventor of the fireworks algorithm (FWA)\cite{tan2010fireworks}, also provided a minimalist version of a fireworks algorithm to make it easier to apply to real-world engineering problems \cite{li2018bare}. These attempts indicate that these optimization methods may have simple and effective structures. Moreover, in 2018, Kennedy made a more abstract proposition that PSO consists of two salient components: a dynamical mechanism governing particle motion and inter-particle communication topology \cite{blackwell2018impact}. This division divided PSO into particle dynamic behavior and information interaction structure, and it further promoted the research of PSO's basic mechanism. This understanding may be considered the prototype of the idea of the modeling of an simplest intelligent optimization system, which has guiding significance for the proposing the quantum dynamics model of a minimalist optimization system.

Hence, in this paper, the research of dynamic model is divided into the dynamic equation and the research of search strategies derived from the approximation of the objective function under this equation. Since some researchers have already used the concept of quantum mechanics to successfully explore topics related to optimization \cite{Nowotniak2010}, the dynamic behavior of the optimization system is studied based on the dynamic equation (Schr\"odinger equation) in quantum mechanics. In 1994, Finnila et al. proposed the quantum annealing algorithm (QAA) by using gradually decreasing quantum fluctuations to obtain a ground-state solution to the Schr\"odinger equation \cite{finnila1994quantum}, in which the optimization problem is transformed as a constraint condition in a quantum system.
In 2015, Mukherjee and Chakrabarti performed a comparative study of quantum annealing techniques and simulated annealing techniques in which they provided the "quantum version" of the Metropolis acceptance criterion \cite{mukherjee2015multivariable}.
In 2005, Das et al. discussed the relaxation dynamics of a system in an energy-constrained system in which spin was retarded by the appearance of energy barriers and followed the kinetic rule \cite{das2005quantum}. Sun assumed that a particle was moving in a delta potential well and proposed quantum-behaved particle swarm optimization (QPSO) \cite{sun2004global} in which a simple potential well was used to approximate the objective function. In these works, the Schr\"odinger equation is often used to establish the relationship between the optimization problem and the quantum potential energy field, which suggests that the Schr\"odinger equation may become a breakthrough in the use of quantum mechanics to study the basic search behaviour of optimization algorithms.

In this work, with the time-dependent Schr\"odinger equation is taken as the equation of quantum dynamics of optimization algorithms and the optimization algorithm is transformed into the ground state wave function and ground state energy problem of constrained states. Then a quantum dynamics model of intelligent optimization algorithm is established, which means the iterative evolution of such a simulated optimization algorithm is governed by the Schr\"odinger equation. Then the evolution of the probability distribution of the optimal solution is similar to the evolution of the wave function in the Schr\"odinger equation. These sampling probes can be regarded as feasible solutions to the black-box optimization problem, and the histogram of those sampling probes can be regarded as the modular square of the wave function. Moreover, to demonstrate the search behaviour under the constraints of the approximated objective function, Taylor expansion is selected to decompose the objective function under different conditions in this work. The basic iterative process (BIP) governed by quantum theory is proposed, which can be seem as a bare-bone algorithm for intelligent optimization under quantum thoery. The experimental results between different bare-bones algorithms confirm the existence and effectiveness of the basic mechanism of quantum dynamics intelligent optimization.
The main contributions of this study can be summarized as follows:

(1) The Schr\"odinger equation is employed as the dynamics equation of the optimization algorithm, which is used to describe the basic iterative process of optimization algorithm.

(2) The intelligent optimization algorithm is decomposed by the Taylor approximation, and the interpretation of the basic mechanism of the intelligent optimization algorithm is provided.

The remainder of this paper is organized as follows. Section~\ref{reason} introduces the reasons why adopt quantum mechanics. Section~\ref{black} introduces intelligent optimization system from quantum perspective. In Section~\ref{sch}, a quantum dynamic model for intelligent optimization is presented and discussed. Section~\ref{theory} provides a detailed elaboration of the quantum dynamic model under different levels of approximation. Evaluation experiments are then presented in section~\ref{Experiments}. Section~\ref{sum} provides some concluding statements.

\section{The reason for choosing quantum mechanics}
\label{reason}

To study the dynamic search mechanism of the optimization algorithm, the most direct method is to find a time evolution equation to describe the optimization algorithm, which is the dynamic equation. Some scholars have begun to realize the connection between the optimization algorithm and the dynamic system \cite{li2019dynamical}. In a broad sense, some classical optimization algorithms are dynamics algorithms. For example, classical genetic algorithm(GA) \cite{mirjalili2019genetic}  can be regarded as a dynamic algorithm designed based on the law of biological evolution, and ant colony optimization (ACO) \cite{dorigo2006ant} and particle swarm optimization (PSO) \cite{schutte2005evaluation} can be regarded as dynamic algorithms that imitate the movement behavior of ants or birds. Simulated annealing (SA) \cite{kirkpatrick1983optimization} is a material thermal dynamics algorithm. In the dynamic searching process, those intelligent optimization algorithm adopts a large number of probabilistic iterative operations, and gives up the requirement of the certainty of solutions, so as to obtain the search ability in a large solution space in an acceptable time.

The deep connection between randomness and intelligent optimization algorithms has long been discovered by computer scientists. In 1976, Turing Prize winner Rabin believed that "what should be abandoned is to obtain results in a completely certain way, which may make mistakes, but the possibility of mistakes is very small, that is to say, probabilistic algorithms can be applied to such problems."  Kennedy, the author of particle swarm optimization, also believes that "the degree of randomness determines the level of intelligence" \cite{eberhart2001swarm}. Stochastic and probabilistic searching mechanisms are widely used in intelligent optimization algorithms to obtain global optimal solutions. The similarity of probabilistic behaviour between optimization algorithms and quantum system suggests that quantum dynamics model is a possible choice for studying the intelligent optimization algorithm with quantum theory.

To propose the quantum dynamical model of an intelligent algorithm, the optimization problem needs to be regarded as the problem of solving the ground state of the constrained quantum system, and the optimization algorithm is governed by the Schr\"odinger equation.

The main reasons for adopting quantum dynamics are as follows:

(1) Through the long-term development of quantum mechanics, a complete mathematical theoretical framework has been established and widely confirmed by the experimental results. Modeling the intelligent optimization system with quantum theory can provide a mature and complete theoretical platform for studying the basic structure of optimization algorithms, and help to analyze and study the basic framework of intelligent optimization algorithms from a deeper perspective.

(2) There are too many kinds of intelligent optimization algorithms and different algorithm models. It is difficult to directly study the generalized dynamical motion behaviours of an intelligent algorithm. Using quantum dynamics to study the intelligent optimization algorithm will limit the motion behavior of the intelligent optimization algorithm to the micro world described by Schr\"odinger equation. Studying the operation law of intelligent optimization algorithm based on the determined equation may provide some theoretical explanations of some common strategies in intelligent optimization algorithm from the perspective of quantum physics.
\section{Intelligent optimization system from the quantum perspective}
\label{black}

\subsection{Global optimization problem}

Numerous scientific problems, such as protein structure prediction, the evolution of fitness landscapes require efficient global optimization of multivariate objective functions \cite{tolkunov2012single}. Hence, the one-dimensional global optimization problem studied in this work can be defined as follows (to simplify, only the global minimum is defined):

\begin{myDef}
A global minimum $\textit{x}_{0} \in \mathbb{X} $ of one function $\textit{f}$ with a limited budget for evaluations $\colon$

$\mathbb{X}\to \mathbb{R} $ is an input element with ${ \textit{f}(\textit{x}_{0})\leq \textit{f}(\textit{x})}$ $\forall \textit{x} \in \mathbb{X}$
\end{myDef}

In optimization, the intelligent optimization algorithm does not have any information such as gradients or the Hessian of a objective function $f(x)$ \cite{golovin2017google}. The information of the objective function can only be obtained through a sampling operation; thus, the intelligent algorithm cannot perform any mathematical transformation on the objective function itself. The solution is uncertain for the optimization system.

\subsection{Probability representation of solutions in optimization}

\begin{figure}[!h]
\centerline{\includegraphics[width=8cm, angle=0]{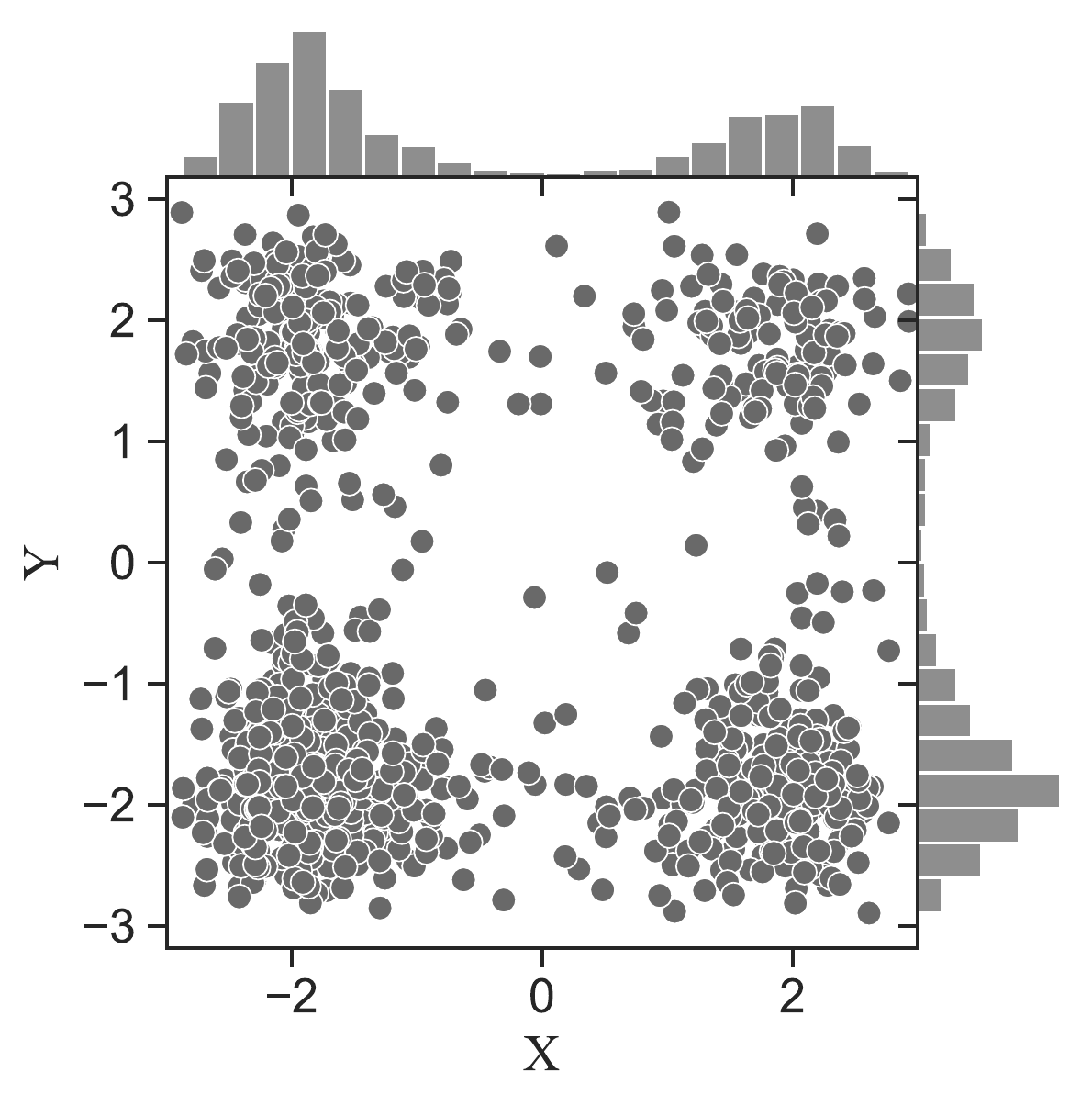} }
\caption{Schematic histogram of the possible solutions on 2D double well potential function}
\label{fig:Noise}
\end{figure}

Since the significant character of a intelligent optimization system is the uncertainty of the solution, in this subsection, to demonstrate the probability characteristics of solution, we start with the two-dimensional double-well potential as an example in Fig.~\ref{fig:Noise}. We assume $V(x)$ to describe two generally asymmetric wells of the form

\begin{equation}
V(x)=\sum_{i=1}^{n} [V_0\frac{ (x_i^2-a^2)^2}{a^4}+\delta\cdot{x_i}]
\label{equ:doublewell}
\end{equation}

where $V_0$, $a$  and $\delta$ ($\delta\geq0$) are real constants. When $\delta=0$, the function has two minima located at $x= \pm{a}$ and separated by a barrier of height $V_0$. If $\delta\cdot{a}\ll V_0$, the two degenerate minima are split by a quantity $\Delta\nu\approx{2}\delta{a}$, and the minimum at $x\approx{-a}$  will be slightly favored.

The goal of the optimization is to find the most likely minimum solutions. The search operations can be viewed as the measurement process affecting the probability that the solution becomes the optimal solution. Since any solution is uncertain, it is very reasonable to express the solution of the optimization problem in terms of probability density $p_k(x)$ of $k$ probes. The expected value of solutions indicates the most possible solution currently.

\begin{myDef}
The expected value of solutions can be defined as follows:
\begin{equation}
\overline{f(x)}=\int f(x)p_k(x)dx
\label{equ:observablee}
\end{equation}
\end{myDef}

When the solution is expressed as a distribution, the optimization system cannot determine the optimal solution theoretically. So, the global optima can also be theoretically defined according to (\ref{equ:zero}) as follows:

\begin{equation}
f_0(x)=\int{f(x)p_k(x)dx}-f_{min}
\label{equ:zero}
\end{equation}
where $p_k(x)$ is the probabilistic distribution of the solutions; $f_{min}$ is the theoretical optimal solution, in Fig.~\ref{fig:Noise}, this can also be represented as a histogram of probes.

It is an important phenomenon that the expected value of solution cannot be zero, which is similar to the zero energy in a quantum system. In quantum systems, the energy of the ground state is not zero but a stable state with finite energy, and it is also strongly wave function relevant. This characteristic of a intelligent optimization system also suggests that the use of quantum mechanics may be a better choice to study the basic search behaviour of optimization algorithms.

\subsection{Optimization process from the quantum perspective}

\begin{figure}[!h]
\centerline{\includegraphics[width=8cm, angle=0]{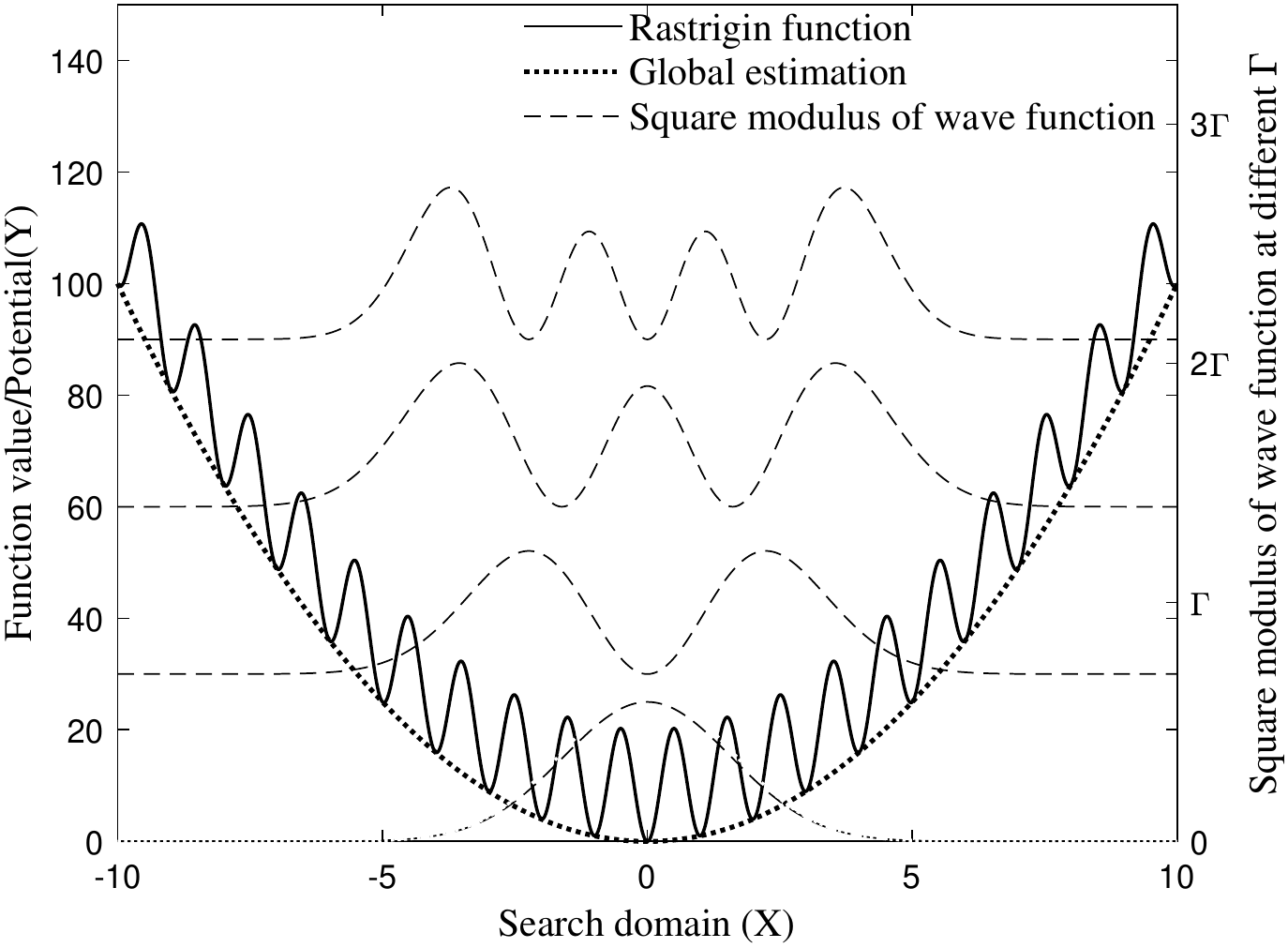} }
\caption{Schematic diagram of large scale well function}
\label{fig:WFBIG}
\end{figure}

Normally, our intuition is that a global negative gradient leads to good solutions. Energy landscapes with such a global gradient easily find the global minimum. As shown in Fig.~\ref{fig:WFBIG}, in the Rastrigin function, $x^2$ is superimposed with small noise $cos(x)$, and the algorithm can find the global optimum only in constant sampling times. However, in realistic problems, the global gradient will be weak and obscured by the local large ``noise" or irregularities in the objective function. Therefore, the global gradient will be invisible to the algorithm at a certain scale. Hence, in a normal optimization system, a multi-scale search process is inevitable.

In quantum theory, when the solution of a optimization problem is regarded as the modulus square of a wave function, it can be proven that the position information and frequency (search scale) of the solution cannot be accurately obtained at the same time. Hence, a multi-scale process is also an inevitable choice.

\begin{figure}[!h]
\centerline{\includegraphics[width=8cm, angle=0]{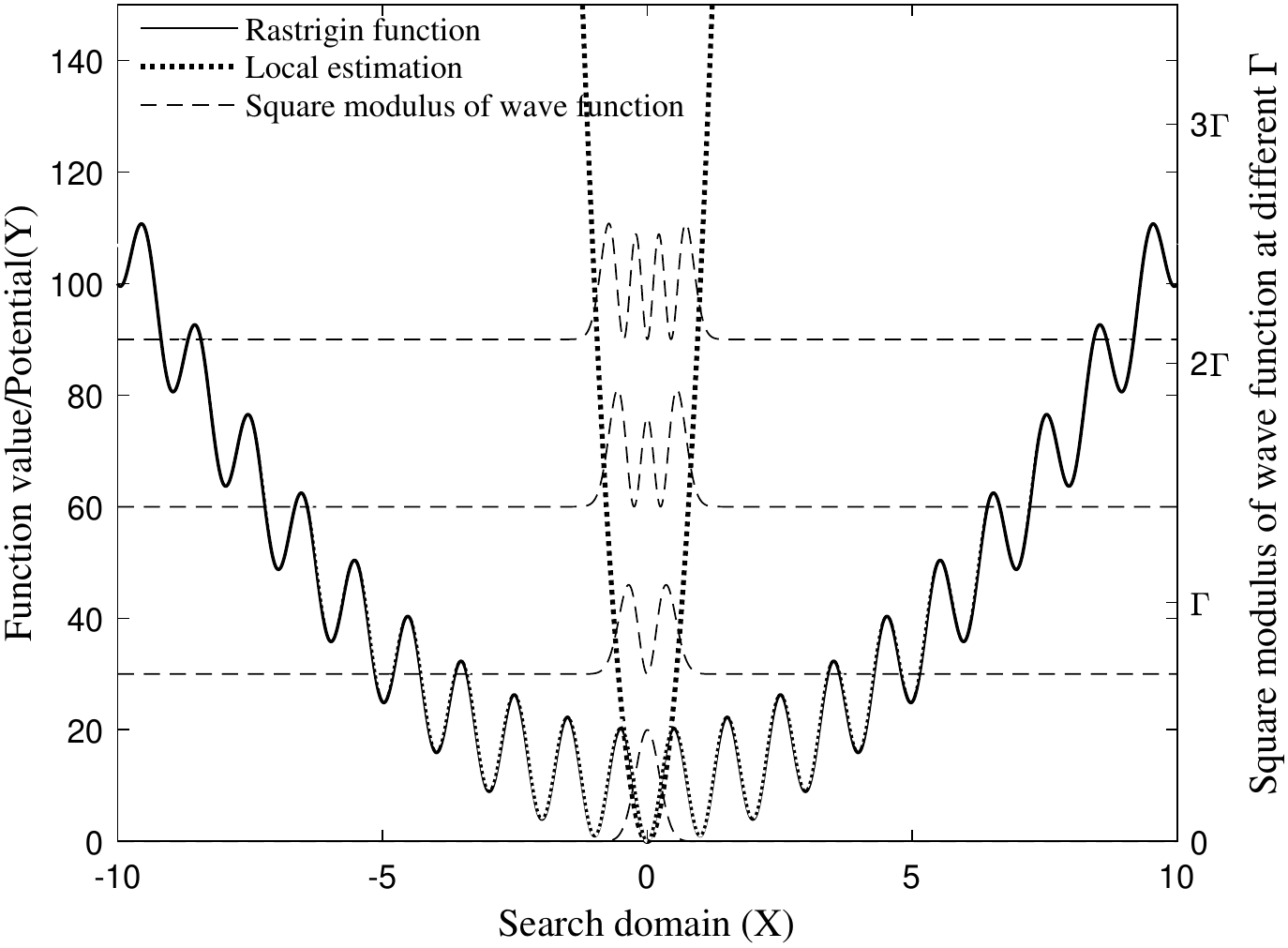} }
\caption{Schematic diagram of small scale well function}
\label{fig:WFSM}
\end{figure}

Moreover, the optimization process is the evolution process of the wave function from a high-energy constraint to a low-energy constraint. As shown in Fig.~\ref{fig:WFBIG}, in a large-scale search, as the constrained energy decreases, the wave function evolves to a ground state, roughly determining the global optimal region. As shown in Fig.~\ref{fig:WFSM}, in a small-scale search, the wave function also evolves from a high-energy constraint to a low-energy constraint, and relatively, a high-precision solution is obtained by using local gradients.


\section{Quantum dynamic model of the intelligent optimization algorithms}
\label{sch}

\begin{figure}[H]
\centerline{\includegraphics[width=24cm, angle=0]{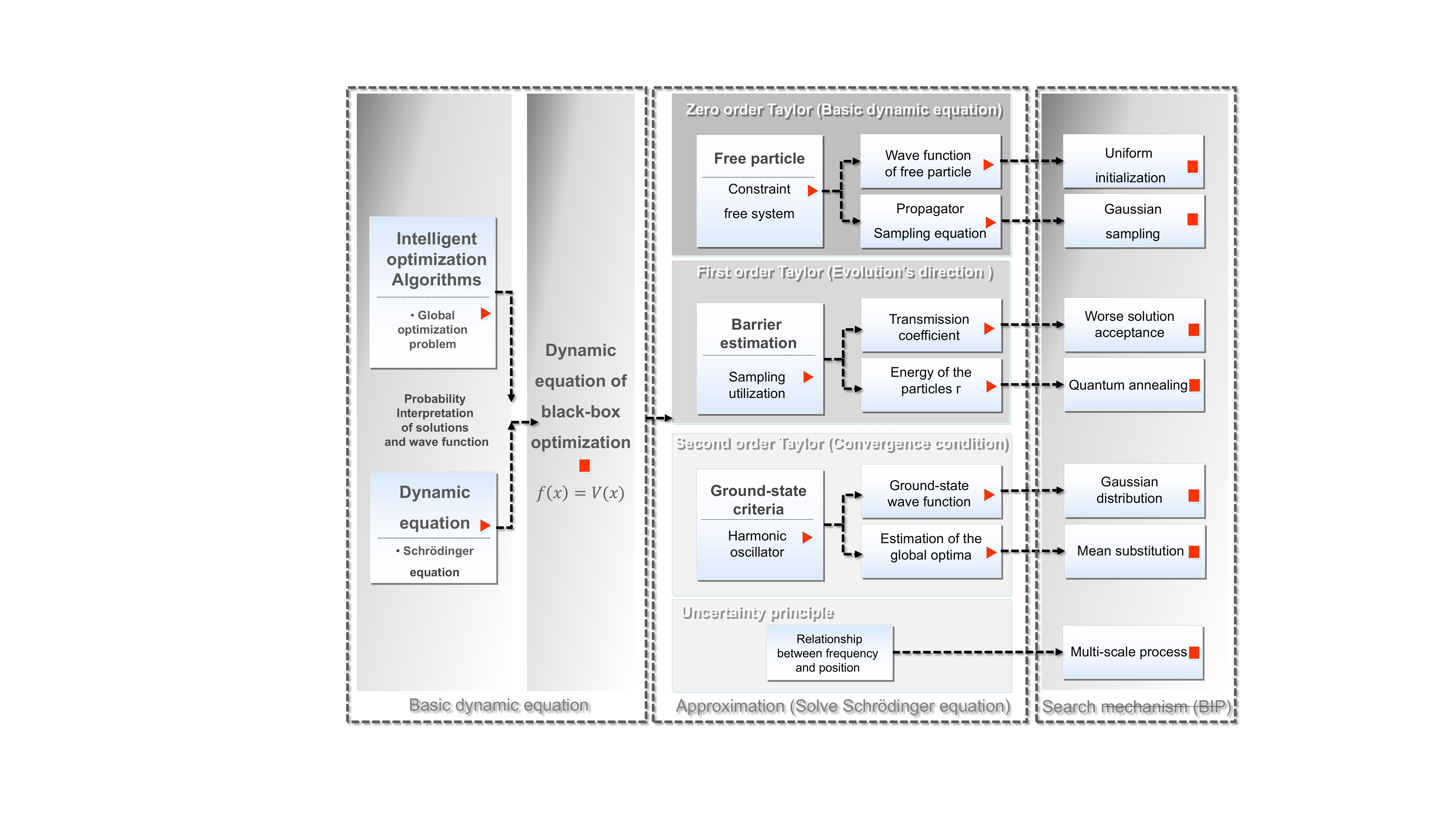} }
\caption{Intelligent optimization algorithm from quantum dynamical perspective}
\label{fig:QModel}
\end{figure}

In this section, the dynamical model of intelligent optimization algorithms is proposed based on the Schr\"odinger equation. With a simple approximation method some basic search strategies are provided, as shown in Fig.~\ref{fig:QModel}. A detailed elaboration is provided in the next section.

\subsection{Basic quantum dynamic equation}

When considering the optimization problem as a potential energy constraint, the dynamic equation of the optimization problem is as follows:

\begin{equation}
\textit{i}\hbar \frac{\partial\Psi(\textit{x},\textit{t})}{\partial \textit{t}} =\left (-\frac{\hbar^2}{2\textit{m}}\frac{\partial^2}{\partial{x}^2}+\textit{f}\left(\textit{x}\right)\right)\Psi(\textit{x},\textit{t})
 \label{equ:Schf}
\end{equation}

To express the optimization problem more generally, by letting $\textit{D}=\frac{\hbar^{2}}{2 \textit{m}}$, the time-dependent Schr\"odinger equation of the optimization problem is rewritten as:

\begin{equation}
\textit{i}\hbar\frac{\partial \Psi(\textit{x},t)}{\partial t}=\left (-D \frac{\partial^{2}}{\partial \textit{x}^{2}}+\textit{f}(\textit{x})\right) \Psi(\textit{x},t)
\label{equ:shortSchf}
\end{equation}

In (\ref{equ:shortSchf}), $\textit{D}$ is a scale parameter that indicates the scale of the optimization problem. The higher the $\textit{D}$ value is, the greater the quantum effect of the system is, and the lower the precision of the solution is, and vice versa.

The modulus of the wave function $|\Psi(\textit{x},\textit{t})|^2$ of the optimization problem, which is the same as the probability interpretation of Born\cite{born2000quantum}, corresponds to the probability distribution of the solutions at time $\textit{t}$ during the optimization process.

The establishment of the dynamic equation for optimization has two important theoretical implications:

(1) The optimization problem is transformed to the problem of the ground-state wave function $\phi_0(\textit{x})$ of the quantum system constrained by potential energy $\textit{f}(\textit{x})$.

(2) In a mathematical sense, the dynamic behaviours of an optimization problem can be described by the Schr\"odinger equation, a linear partial differential dynamic equation.

It is convenience to study the search mechanism of the optimization system with quantum theory methods.

\subsection{Approximation of the optimization problem}

For studying the basic behaviour of the corresponding optimization algorithms, it is necessary to take the optimization problem can be decomposable by some approximation methods, then the dynamic behaviour of the algorithms under some simple structures are able to be studied.

In this paper, the Taylor expansion is employed to simplify the objective function given by Eq.~(\ref{equ:Schf}). Without losing generality, this paper only studies single objective function, and multi-objective optimization can be treated in the same way. The objective function $\textit{f}(\textit{x})$ is first assumed to be continuously differentiable and is expanded as follows:

\begin{equation}
\textit{f}(\textit{x})= \sum _{\textit{n}=0}^{\infty} \frac{\textit{f}^{\textit{n}}(\textit{x}_0)}{\textit{n}!}(\textit{x}-\textit{x}_0)^\textit{n},
\label{equ:taylorexpasion}
\end{equation}
where $\textit{n}!$ represents the factorial of $\textit{n}$ and $\textit{f}^{\textit{n}}(\textit{x}_0)$ is the $\textit{n}$th derivative of $\textit{f}$ at point $\textit{x}_0$.

According to the different Taylor approximations of the potential energy function, the optimization algorithm can determine how to use the sampling information to propose the corresponding search behaviours. The significance of decomposition of the optimization problem to the optimization system can be classified as follows:

(1) The zero-order Taylor approximation of $\textit{f}(\textit{x})$ is $\textit{f}^{(0)}(\textit{x})=\textit{f}(\textit{x}_0)$, and $\textit{f}(\textit{x}_0)$ is a constant, so by translation, $\textit{f}^{(0)}(\textit{x})=0$. The relative concept of quantum physics is the wave function and propagator of free particles.

(2) The first-order Taylor approximation of $\textit{f}(\textit{x})$ is $\textit{f}^{(1)}(\textit{x})=\textit{f}^{(1)}(\textit{x}_0)(\textit{x}-\textit{x}_0)$; if the point $\textit{x}_0$ is an extreme point, then its first derivative $\textit{f}^{(1)}(\textit{x}_0)= 0$, so the first-order Taylor approximation near the extreme point is $\textit{f}^{(1)}(\textit{x})=0$; thus, the relative quantum physics concepts are quantum tunneling and quantum annealing.

(3) The second-order Taylor expansions are $\textit{f}^{(2)}(\textit{x})=\frac{1}{2} f^{(2)}\left(x_{0}\right)\left(x-x_{0}\right)^{2}$ by assuming that the higher-order terms are negligible when $(\textit{x}-\textit{x}_0)$ is sufficiently small. The relative concept of quantum physics is the ground state wave function of a quantum harmonic oscillator.

\subsection{Algorithm based on basic dynamic behaviours}

Based on previous elaboration, we preliminarily believe that there are some basic structures of the optimization algorithm based on the quantum dynamics equation as follows:

(1) The zero-order Taylor approximation of the objective function is the basic dynamic equation, which explains the random normal sampling search behaviour of the optimization algorithm.

(2) The first-order Taylor approximation of the objective function can be seen as the use of slope information of any two samplings of the objective function, which explains the mechanism of obtaining the gradient information of the objective function in the sampling process.

(3) The second-order Taylor approximation of the objective function is a quantum harmonic oscillator. The normal distribution can be used as the ground-state convergence criterion in the iterative process of the algorithm under the second-order approximation.

(4) The uncertainty principle of the optimization algorithm indicates the quantum nature and necessity of the multi-scale sampling process of the optimization algorithm \cite{Wangxin2020}.
\section{Quantum dynamic behaviours of the intelligent optimization algorithms}
\label{theory}

\subsection{Basic dynamic equation under zero-order Taylor approximation}

The zero-order Taylor approximation of the objective function $\textit{f}(\textit{x})$ is approximate to a constant, which does not contain gradient information. The global optimal value of the function can be shifted to zero. Therefore, under the zero-order Taylor approximation, the objective function can be regarded as a constant equal to zero as:

\begin{equation}\textit{f}(\textit{x})=0\end{equation}

So the dynamic equation of the optimization problem (\ref{equ:shortSchf}) can be recast as:

\begin{equation}
\textit{i}\hbar \frac{\partial \Psi(\textit{x},t)}{\partial t}=-\textit{D} \frac{\partial^{2} \Psi(\textit{x},t)}{\partial \textit{x}^{2}}
\label{equ:schOP}
\end{equation}
$\textit{D}$ is called scale coefficient. That dynamic equation satisfies the free particles system in quantum mechanics.

Since then, a propagator of the free particle in quantum mechanics and quantum field theory is the probability amplitude that describes the particle moving from one place to another at a specific time \cite{Griffiths2006}. The propagator is also a Green function of the equation of motion of the field as follows:

\begin{equation}
\begin{aligned}
K\left(x, x^{\prime} ; t\right) &=\frac{1}{2 \pi} \int_{-\infty}^{+\infty} d k e^{i k\left(x-x^{\prime}\right)} e^{-i \hbar k^{2} t /(2 m)}  \\
& \left(\frac{m}{2 \pi i \hbar t}\right)^{1 / 2} e^{-m\left(x-x^{\prime}\right)^{2} /(2 i \hbar t)}
\end{aligned}
\end{equation}

According to quantum field theory, the probability of appearing at position $\textit{x}$ after $\Delta \textit{t}$ can be expressed by Green's function:

\begin{equation}
\textit{G}\left(\textit{x}, \textit{x}^{\prime}, \Delta \textit{t}\right)=\frac{1}{\sqrt{4 \pi iD\hbar \Delta t}} \textit{e}^{-\frac{\left(\textit{x}-\textit{x}^{\prime}\right)^{2}}{4 iD\hbar \Delta \textit{t}}}
\label{equ:greenfunction}
\end{equation}

By letting $\sigma=\sqrt{2iD\hbar \Delta t}$, (\ref{equ:greenfunction}) can be recast as:

\begin{equation}
\textit{G}\left(\textit{x}, \textit{x}^{\prime}, \Delta \textit{t}\right)=\frac{1}{\sqrt{2 \pi} \sigma} \textit{e}^{-\frac{\left(\textit{x}-\textit{x}^{\prime}\right)^{2}}{2 \sigma^{2}}}
\label{equ:normGreenfunction}
\end{equation}
where Green's function of the Schr\"odinger equation is transformed to a normal distribution, where $\sigma$ is the scale parameter of the sample. The size $\sigma$ is related to the scale coefficient $\textit{D}$ and the time interval $\Delta \textit{t}$ of the Schr\"odinger equation. The Green function $G(x,x',\Delta t)$ describes the Markov process of sampling. Let the particle's position at time $\textit{t}$ be $\textit{x}(\textit{t})$; then, the position $\textit{x}(\textit{t} + \Delta \textit{t})$ is:

\begin{equation}
\textit{x}(\textit{t}+\Delta \textit{t})=\textit{x}(\textit{t})+\sigma \textit{N}(0,1)
\label{equ:Greenmove}
\end{equation}
where $\textit{N}(0,1)$ is the standard normal distribution and $\sigma$ is the sampling scale, which continuously decreases when the time interval $\Delta \textit{t}$ decreases (frequency information increase) and can gradually improve the accuracy of the search. This corresponds to the multi-scale sampling behaviour, which exists in the sampling process of a large number of optimization algorithms. 
After a certain number of iterations, the distribution of these sampling individuals in the domain is the approximation of the modulus of the wave function $|\Psi(\textit{x})|^2$ after the evolution.

In general, under the zero-order Taylor approximation of intelligent optimization problem, the particle in quantum space is unconstrained, and its sampling behaviour is a random normal walk of a large number of sampled individuals.

\subsection{Potential barrier tunneling under first-order Taylor approximation}

To further decompose the basic operation of the optimization algorithm, the first-order Taylor approximation is employed for any point $x_0$ in the domain of definition in the quantum dynamics equation of the optimization algorithm. Point $x_0$ can be regarded as the sampling point of a current sampling individual. Under the condition of not affecting the position of the optimal solution, the objective function value is shifted. Letting $f(x_0)=0$, the Schr\"odinger equation is recast as follow:

\begin{equation}
i \hbar \frac{\partial \psi(x, t)}{\partial t}=\left[-D \frac{\partial^{2}}{\partial x^{2}}+\left(\frac{\partial f\left(x_{0}\right)}{\partial x}\left(x-x_{0}\right)\right)\right] \psi(x, t)
\label{equ:first}
\end{equation}
where, $\Psi(x, t)$ is the probability distribution of the optimal solution at $t$ in the iterative process of optimization algorithm.

In the above formula, there is more than one term $\frac{\partial f\left(x_{0}\right)}{\partial x}\left(x-x_{0}\right)$ compared with the zero-order Taylor expansion, in which $x_0$ is the current sampling location and $x$ is the next new sampling location. This item shows theoretically that the optimization algorithm needs to perform a derivative operation on the objective function in the iterative evolution process to obtain the current slope $\frac{\partial f\left(x_{0}\right)}{\partial x}$ of the objective function. The gradient information of the objective function near $\textit{x}_0$ can be obtained by first-order Taylor approximation, and its expansion is as follows:

\begin{equation}
\frac{\partial f\left(x_{0}\right)}{\partial x}\left(x-x_{0}\right) \approx \frac{f(x)-f(x_{0})}{\textit{x}-\textit{x}_{0}}\left(\textit{x}-\textit{x}_{0}\right)
\end{equation}

Letting $\Delta f=f(x^{\prime})-f(x_{0})$, equation~(\ref{equ:first}) is recast as follows:

\begin{equation}
i \hbar \frac{\partial \psi(x, t)}{\partial t}=\left[-D \frac{\partial^{2}}{\partial x^{2}}+ f(x^{\prime})-f(x_{0})\right] \psi(x, t)
\end{equation}

This equation describes the quantum dynamic process of the optimization algorithm in two samplings. Since the time difference $\Delta t=t^{\prime}-t_0$ between the two samples is very small and $f(x^{\prime})$ is irrelevant to time $t$, the relationship between the wave function $\psi(x)$ and the position $x$ at $t^{\prime}$ can be established by solving the stationary Schr\"odinger equation as follows:

\begin{equation}
-D \frac{\partial^{2} \psi(x)}{\partial x^{2}}+f(x^{\prime}) \psi(x)= f(x_{0})\psi(x)
\label{equ:staSchor}
\end{equation}
where $f(x_{0})$ is the current potential energy of the particle, and $f(x^{\prime})$ is the sampling result of the potential field.

As shown is Fig.2.,when $\Delta f=f\left(x^{\prime}\right)-f\left(x_{0}\right) \leq 0$, a better solution is obtained by normal sampling from point $x_0$. In this case, there is no barrier between the two samples. According to the first-order Taylor approximation of the quantum dynamics equation, the new position is obtained and directly received as a new solution instead of $x_0$.

When $\Delta f=f\left(x^{\prime}\right)-f\left(x_{0}\right) \textgreater 0$, a worse solution is obtained by normal sampling from point $x_0$. In this case, the particles encounter the barrier when they move randomly. In quantum theory, if a particle can cross the barrier through the tunneling effect, the barrier can be estimated with the barrier height is $\Delta f$ and the barrier width $\left|x^{\prime}-x_{0}\right|$ as follows:

\begin{equation}
V(x)=\left\{\begin{array}{l}
\Delta f, 0<x<\left|x^{\prime}-x_{0}\right| \\
0, x>\left|x^{\prime}-x_{0}\right|, x<0
\end{array}\right.
\end{equation}
where the probability of tunneling through the barrier is the receiving probability of the solution. Letting $k^{\prime}=\sqrt{\frac{1}{D}(\Delta f-E))}$, the solution of (\ref{equ:staSchor}) can be recast as follows:

\begin{equation}
\psi(x)=C_1 e^{-k^{\prime} x}+C_2 e^{k^{\prime} x}
\end{equation}
since $\psi(x)$ must be finite, when $x\rightarrow\infty$, so $C_2=0$

\begin{equation}
\left|\psi(x)\right|^{2} = e^{-2x \sqrt{\frac{1}{D}(\Delta f-E)} }
\end{equation}

\begin{figure}[!h]
\centerline{\includegraphics[width=8cm, angle=0]{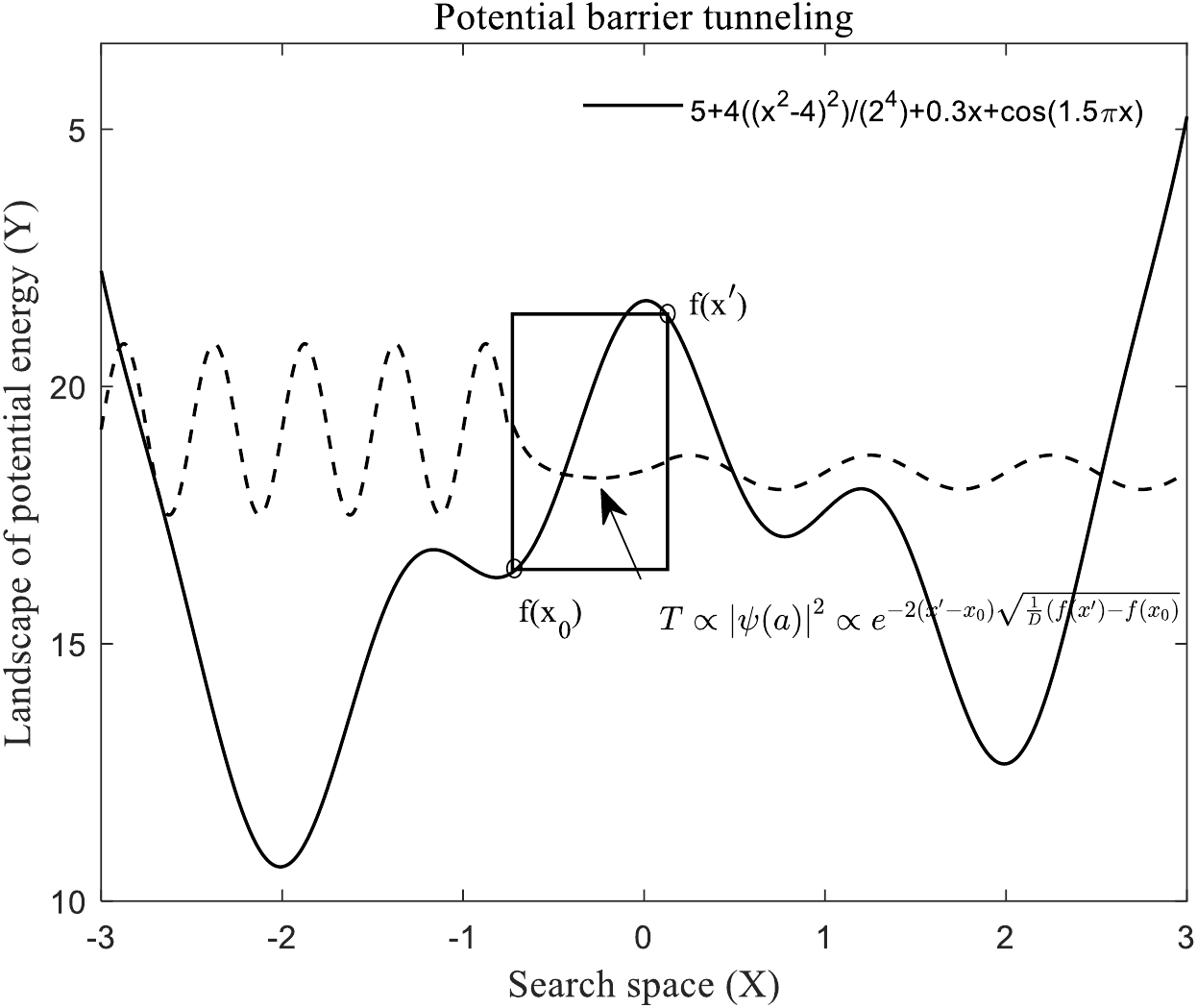} }
\caption{schematic diagram of sin function superimposed with weak random interference}
\label{fig:MQOMODEL}
\end{figure}

According to the barrier penetration principle, the receiving probability is as follows:
\begin{equation}
T \propto\left|\psi(x^{\prime})\right|^{2} \propto e^{-2 (x^{\prime}-x_0) \sqrt{\frac{1}{D}(\Delta f-E)}}
\label{equ:acceptrate}
\end{equation}

According to Mukherjee's work \cite{mukherjee2015multivariable}, the physical constants are reduced, and the barrier height $\Delta f$ scales as $N$ while the kinetic energy $E$ scales as $N\Gamma$; then, the transmission rate is recast as follows:

\begin{equation}
T \propto Ae^{- \Delta x \sqrt{N-N\Gamma}} \sim Ae^{\frac{- \Delta x \sqrt{\Delta f}}{\Gamma}}
\label{equ:acceptrate2}
\end{equation}
where $\Delta x=x^{\prime}-x_0$, according to the concept of quantum annealing in Arnab Das's work \cite{das2005quantum}, at each scale $\sigma_s$, the attenuation is performed according to the following formula:

\begin{equation}
\Gamma =  \Gamma_0 e^{\frac{-1}{Ac}}
\label{equ:annealingC}
\end{equation}
where $\Gamma_0$ can be set as the current search scale $\sigma_s$ and $Ac$ is the annealing coefficient. The above sampling criterion is called the barrier penetration criterion of the optimization algorithm, which is an iterative operation obtained under the first-order Taylor approximation of the objective function.

\subsection{Convergence conditions under second-order Taylor approximation}

The second-order Taylor expansion in the quantum dynamics equation near the extreme point $x_0$ of the objective function is as follows:

\begin{equation}
i\hbar \frac{\partial \psi(x, t)}{\partial t}=-\left[D \frac{\partial^{2}}{\partial x^{2}}-\frac{1}{2} \textit{f}^{(2)}\left(x_{0}\right)\left(x-x_{0}\right)^{2}\right] \psi(x, t)
\label{equ:secondf}
\end{equation}

Equation~(\ref{equ:secondf}) is the same as Schr\"odinger equation of harmonic oscillator in quantum theory. In quantum theory, the complex vibration of the equilibrium position and Lennard-Jones potential are approximated by the harmonic oscillator potential.

In the intelligent optimization, the global optimal position $\textit{x}_0$ in the equation is unknown, so we cannot directly solve the equation. However, the ground-state wave function of a quantum system with a harmonic oscillator potential energy constraint is a normal distribution. Therefore, the ground-state wave function corresponding to the second-order approximation is also a normal distribution. So the module square of the ground-state wave function of the harmonic oscillator can be employed to represent the second-order components of the solution's distributions of the optimization problem.

Normally, the ground-state wave function corresponding to the complex objective function is still complex. However, in the global optimization problem, what we are concerned about is the probability distribution of possible solutions near the optimal solution. Therefore, the normal distribution can be used as the objective criterion for the population to converge to the ground state at a certain scale in the iterative process.

In this case, the theoretical mean value of solution positions is $\bar{x}=\int x\left|\psi_{0}(x)\right|^{2} d x$, which indicates that the expectation of the solution can be obtained by the mean of the sample individual position in the group when the ground state is a normal distribution. Therefore, the worst solution is replaced with the mean value of the sampling position to accelerate the convergence speed to ground-state \cite{ye2019impact}. It shows that the normal criterion and the mean replacement operation are the common operations of the optimization algorithm when the algorithm converges to the ground state.

\subsection{Pseudocode of Basic Iteration Process}

The Basic Iteration Process (BIP) can be obtained according to the basic iteration operation of the optimization algorithm under the quantum dynamics model. BIP is a multi-scale normal population sampling process composed of k independent sampling individuals. The positions of k samples are randomly initialized in the definition domain. Then k samples are independently sampled in the normal sampling iteration. The initial standard deviation $\sigma$ of normal sampling can be larger, for example, the size of the definition domain. During the sampling process, the barrier penetration criterion is implemented, which is directly accepted for the optimal solution and the probability of transmission coefficient is accepted for the worst solution. If the standard deviation of k sampling individuals is less than the current normal sampling standard deviation $\sigma$, the algorithm is considered to have reached the ground state of this scale. Then, the worst solution is replaced by the mean position of K sampling individuals, and the normal sampling standard deviation of all samples is halved to enter the next iteration process of smaller scale. Finally, the algorithm ends when it meets the maximum number of iterations or a certain precision. The BIP under quantum dynamics model is proposed in Algorithm \ref{tab:pseudocode}.

\begin{algorithm}[!htb]
\caption{Pseudocode of BIP}
\label{tab:pseudocode}
     Randomly generate $k$ copies of free particle in the domain [UB,LB], $\sigma_s$=UB-LB \\
        \While {(stop condition is not satisfied)}
          { Initialize $Ac=0$, $\Gamma_0=\sigma_s$ \\
          \While { ($\sigma_k<\sigma_s$)}
       {
                \For{$i$=1 to $k$}{
                Generate $x_i^{\prime}$ for $x_i$ to Eq.~(\ref{equ:Greenmove})  \\

            \uIf{(  $f(x_i^{\prime})<f(x_i)$)}{
             $x_i=x_i^{\prime}$, update the $i$th particle\\
              }
              \Else{
              $x_i=x_i^{\prime}$, update the $i$th particle according to $T$ in Eq.~(\ref{equ:acceptrate2}) \\
              }
              }
              $Ac=Ac+1$ \\
              Calculate the $\sigma_k$ for $k$ copies\\
              }
              update the $\Gamma$ according Eq.~(\ref{equ:annealingC}) \\
              $x^{worst}$=$x^{mean}$ \\
              $\sigma_s$=$\sigma_s/2$
       }
  output
\end{algorithm}

In summary, the optimization algorithm includes three basic strategies: (1) a normal distribution sampling strategy; (2) a oriented particle movement strategy; and (3) a multi-scale strategy. It is found that these three strategies and their variants are used in most optimization algorithms but in different forms.

In the intelligent optimization, the sampling behaviour determines the basic search ability of the algorithm. On this basis, in order to obtain better performance, the algorithm introduces a priori knowledge based on the assumption of the objective problem, so as to orient the movement direction of particle. Usually the multi-scale process is necessary to obtain the structure of the objective problem in complex conditions.
\section{Experiments and discussion}
\label{Experiments}

In the experimental section, the basic search behaviour of BIP is analyzed with the paraboloid function and double well function and some selected basic benchmark function. The proposed method is compared with three representative bare-bone algorithms to explore the similarity of bare-bone schemes under different algorithm frameworks through experiments. The purpose of introducing the other three bare-bone algorithms here is not to compare the performance but to make a reference to the basic performance of the basic model and to evaluate the intelligent optimization model of quantum dynamics as objectively as possible.

\subsection{Benchmark functions}

In this section, experiments were conducted using a 2D double well function, a 2D paraboloid function and the 12 selected uni- and multimodal single-objective optimization benchmark suites \cite{liang2013problem}, including six multimodal functions (F1-F6) and six uni-modal functions (F7-F12) shown in Table~\ref{tab:benchmark}. The later 12 functions represent 12 basic minimization problems with a minimum value zero.

\begin{table*}[ht]

\scriptsize
\setlength{\tabcolsep}{3.5pt}
\centering
\caption{12 benchmark test functions (U: unimodal functions; M: multimodal functions)}
\renewcommand{\multirowsetup}{\centering}
\begin{tabular}{p{6em}m{8cm}cccc} 
\hline
Function Name&{\makecell {Benchmark Function} }&Type&D&Search Space&Global Optimum\\\hline

Griewank &$F_{1}$=$\frac{1}{4000} \sum_{i=1}^{n} x_{i}^{2}-\prod_{i=1}^{n} \cos \left(\frac{x_{i}}{\sqrt{i}}\right)+1$&M&n&[-100,100]&$f(0,...,0)=0$ \\

Rastrigin&$F_{2}=10n+\sum_{i=1}^{n}[x_i^2-10\cos (2\pi x_i)]$&M&n&[-5.12,5.12]&$f (0,...,0)=0$\\

{Ackley}&{$F_3=-20exp(-0.2\sqrt{\frac{1}{n}\sum_{i=1}^{n} x_i^2})-exp(\frac{1}{n}\sum_{i=1}^{n}cos(2\pi x_i))+20+e
$}&M&{n}&{$[-32.77,32.77]$}&{$f (0,...,0)=0$}\\

Levy&$F_{4}=sin^2 (\pi\omega_1)+\sum_{i=1}^{n-1}(\omega_i-1)^2[1+10sin^2 (\pi\omega_i+1)]+ (\omega_n-1)^2[1+sin^2 (2\pi\omega_n)]$,where $\omega_i=1+\frac{x_i-1}{4}$,for all $\quad i=1,...,n$&M&n&[-10,10]&$f (1,...,1)=0$\\

Alpine&$F_{5}=\sum_{i=0}^{n-1}|x_i sin(x_i)+0.1x_i|$&M&n&[0,10]&$f(0,...,0)=0$\\

Schwefel&$F_{6}=418.9829n-\sum_{i=1}^{n}x_i\sin(\sqrt{|x_i|})$&M&n&[-500,500]&$f(420.97...)=0$\\

Sphere&$F_{7}=\sum_{i=1}^nx_i^2$&U&n&[-5.12,5.12]&$f (0,...,0)=0$ \\

Sum Squares&$F_{8}=\sum_{i=0}^{n-1}i x_i^2$&U&n&[-10,10]&$f(0,...,0)=0$\\

Rotated Hyper-Ellipsoid&\multirow{3}{*}{$F_{9}=\sum_{i=1}^{n}(\sum_{j=1}^{i}x_i)^2 $}&\multirow{3}{*}{U}&\multirow{3}{*}{n}&\multirow{3}{*}{[-65.54,65.54]}&\multirow{3}{*}{$f(0,...,0)=0$} \\

Ellipsoidal&$F_{10}=\sum_{i=1}^{n}(x_i-i)^2$&U&n&[-100,100]&$ (1,2,...n)=0$ \\

Sum of Different Power&\multirow{2}{*}{$F_{11}=\sum_{i=1}^{n}(|x_i|^{i+1} $}&\multirow{2}{*}{U}&\multirow{2}{*}{n}&\multirow{2}{*}{[-1,1]}&\multirow{2}{*}{$f(0,...,0)=0$} \\

Zakharov&$F_{12}=\sum_{i=1}^{n}x_i^2+(\sum_{i=1}^{n}0.5ix_i)^2+(\sum_{i=1}^{n}0.5ix_i)^4$&U&n&[-5,10]&$f(0,...,0)=0$\\

\hline
\label{tab:benchmark}
\end{tabular}
\end{table*}

\subsection{Experimental settings}

The results of each experiment were obtained based on 51 independent trial runs, and the maximum number of function evaluations ($\textit{MaxFES}$) in each run was set according to different experimental requirements, where $D$ is the dimension of the problem. The error values $f(x)-f(x^{\ast})$ were stored for each trial run. The parameters of the schemes used for comparison purposes are set according to the values provided in the literature, which are listed in Table~\ref{tab:parameters}.

\begin{table}[]
\centering
\scriptsize
\setlength{\tabcolsep}{3.5pt}
\renewcommand\arraystretch{1.5}
\caption{Parameter settings of optimization algorithms}
\begin{tabular}{cl}
\hline
Algorithm & Parameter settings                                                                                                                    \\ \hline
BIP  &  NP=15\\
BBPSO       & NP=20 \\
BBFWA   & NP=300  \\
GBDE      & NP=100, $C_r$=$N(0.5,0.1)$                                                                                                  \\
\hline
\end{tabular}
\label{tab:parameters}
\end{table}

\subsection{Investigation of evolutionary process of BIP}

\begin{figure}[H]
\centering

\subfigure[\tiny{Double well function (FE=1) }]{\includegraphics[width=4.5cm,angle=0]{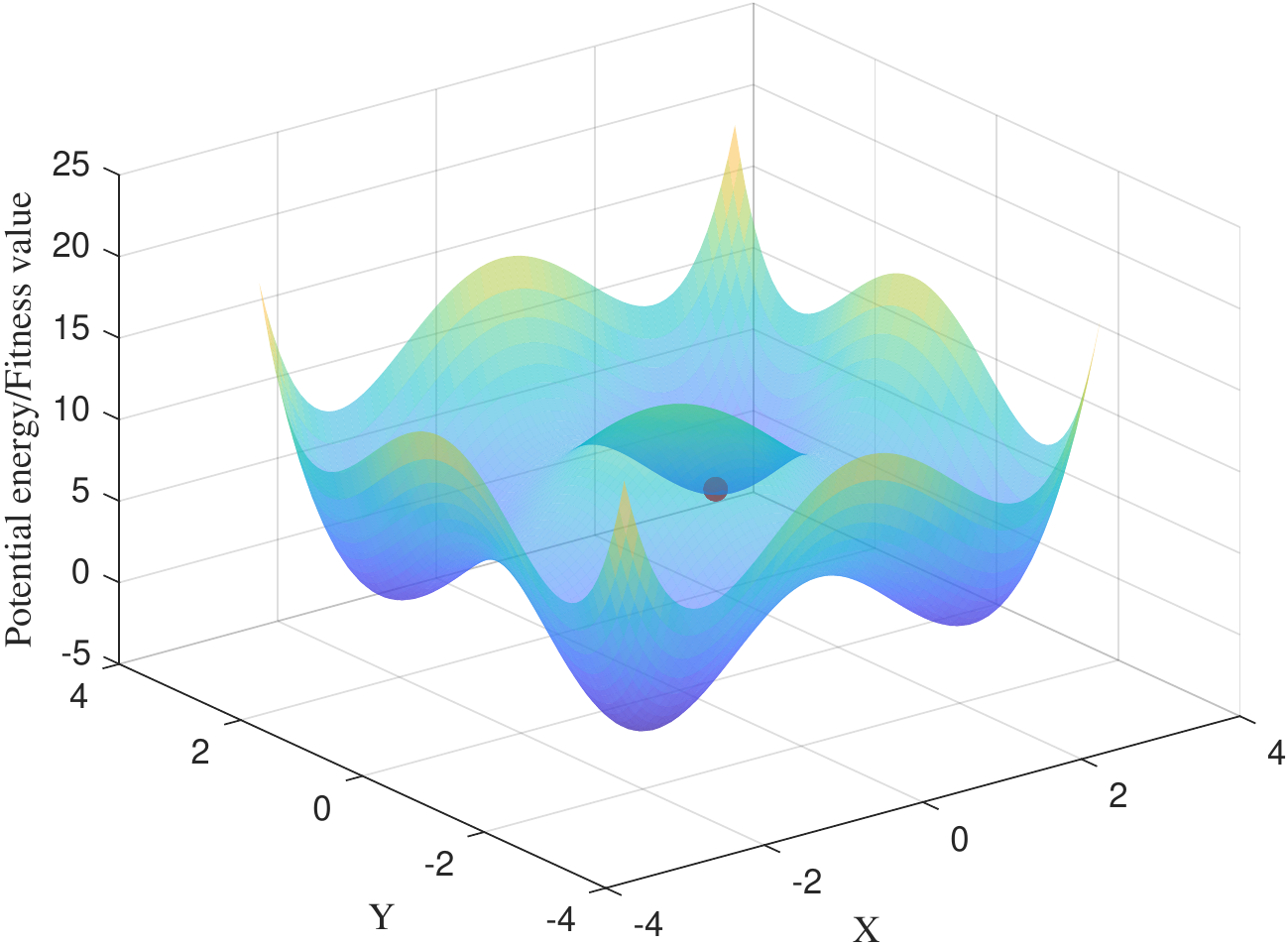} }
\subfigure[\tiny{Particle trajectories (without second-order estimation FE=200)}]{\includegraphics[width=4.5cm,angle=0]{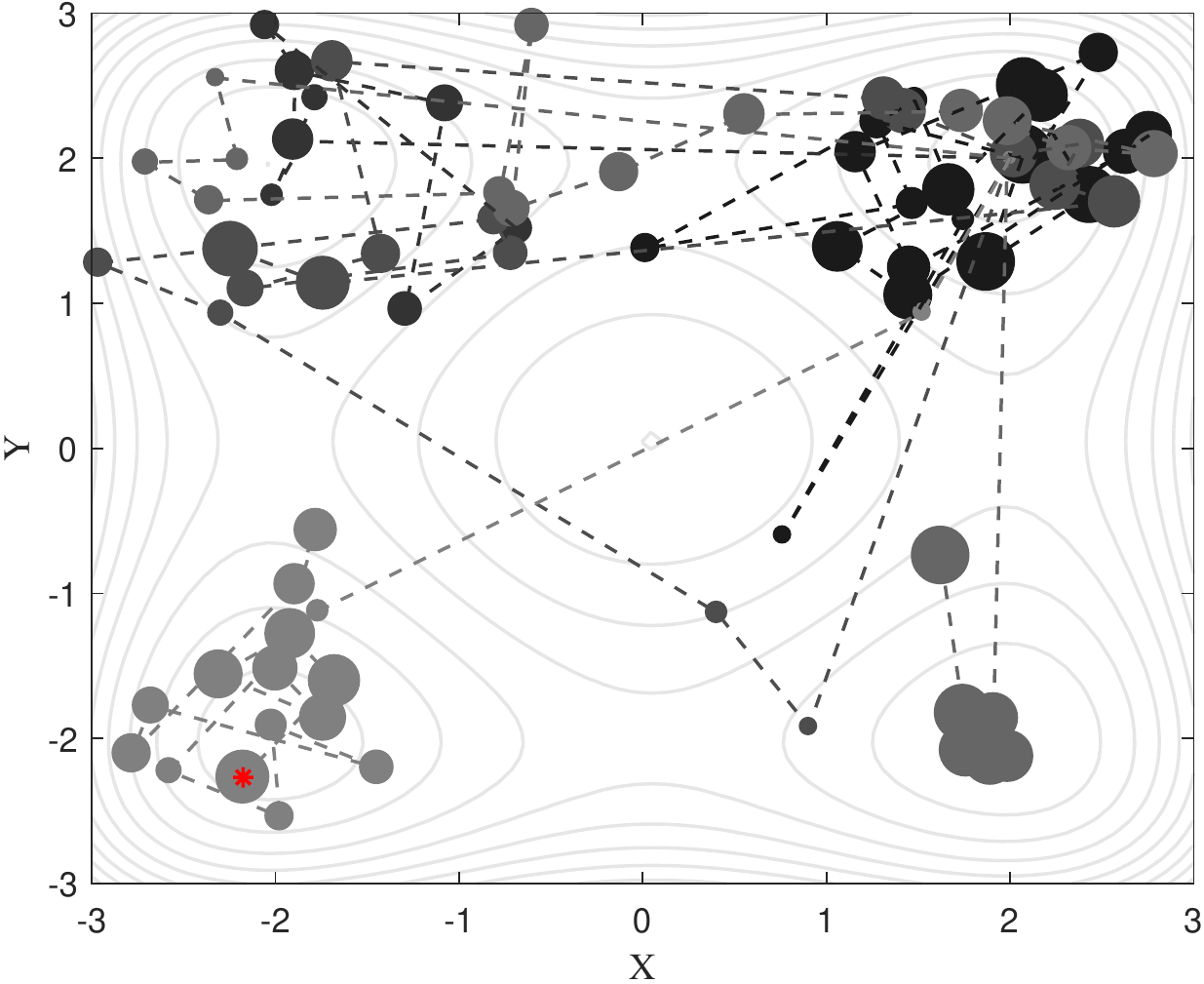} }
\subfigure[\tiny{Modulus of wave function $|\psi(x)|^{2}$ (FE=200)}]{\includegraphics[width=4.5cm,angle=0]{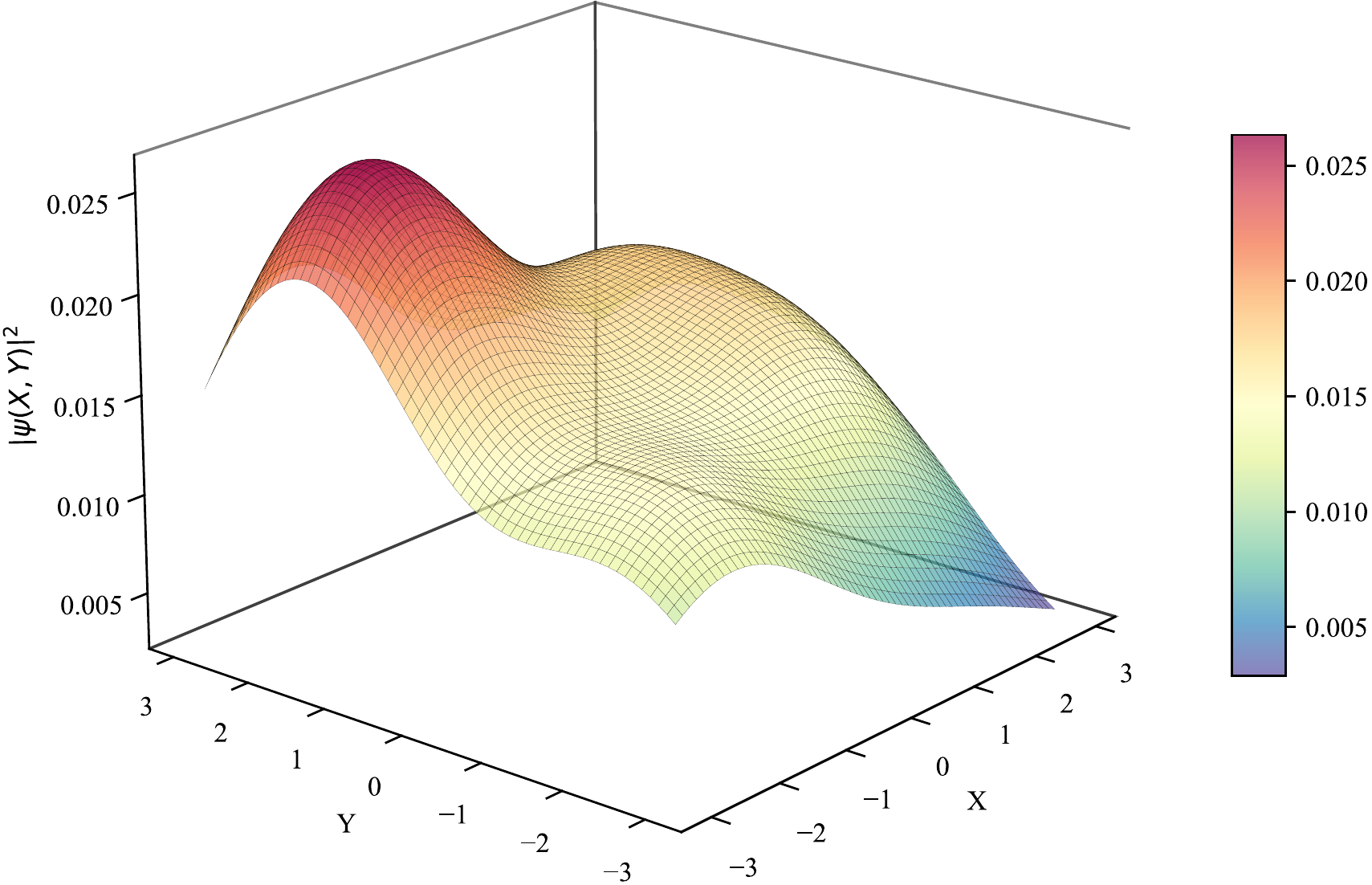} }

\subfigure[\tiny{Double well function (FE=1)}]{\includegraphics[width=4.5cm,angle=0]{DW_Initial.pdf} }
\subfigure[\tiny{Particle trajectories (with second-order estimation FE=200)}]{\includegraphics[width=4.5cm,angle=0]{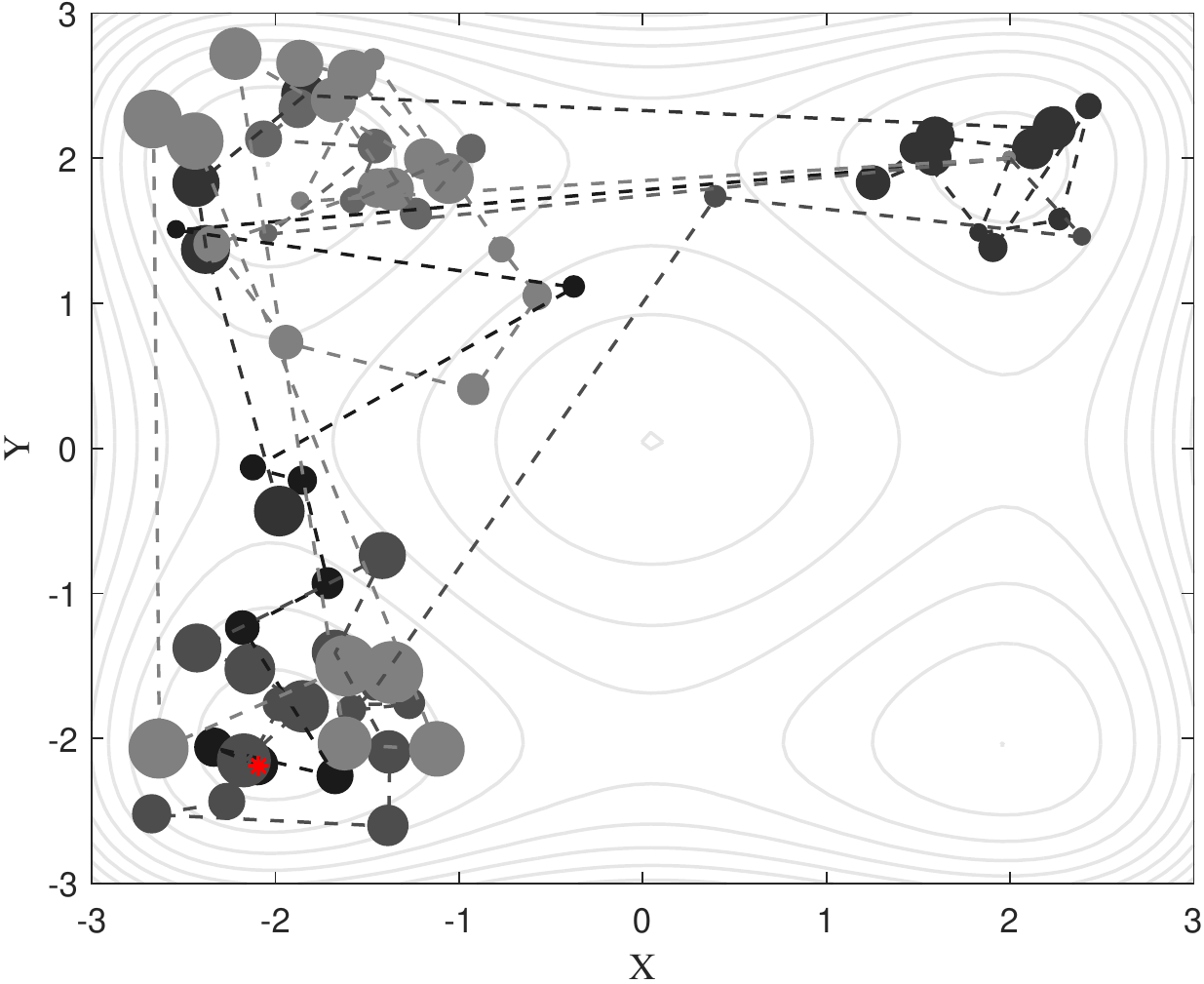} }
\subfigure[\tiny{Modulus of wave function $|\psi(x)|^{2}$ (FE=200)}]{\includegraphics[width=4.5cm,angle=0]{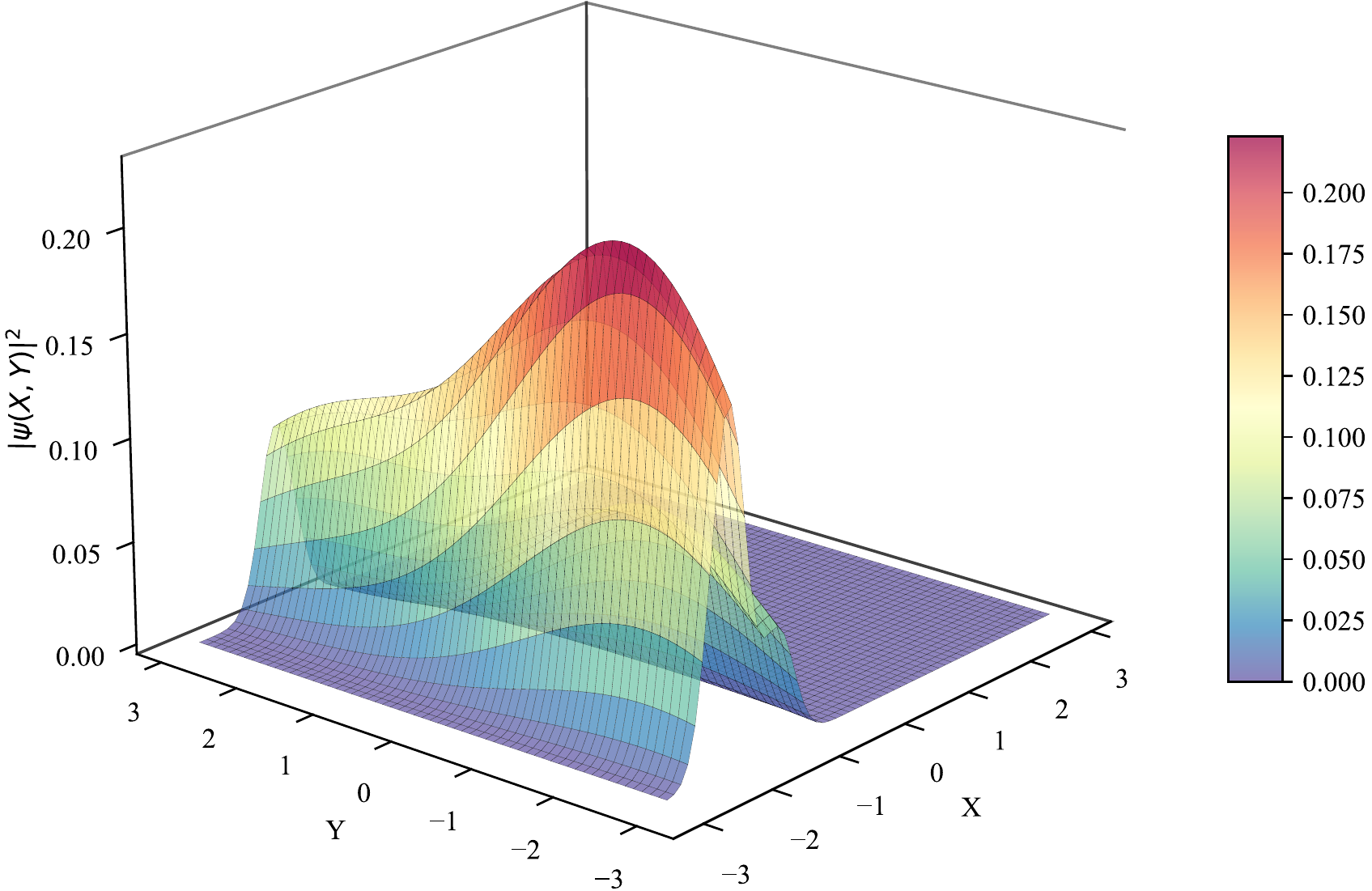} }
\caption{Evolution of algorithm under the different approximation of the double well function}
\label{fig:DWtrack}
\end{figure}

\begin{figure}[H]
\centering

\subfigure[\tiny{Paraboloid function (FE=1) }]{\includegraphics[width=4.5cm,angle=0]{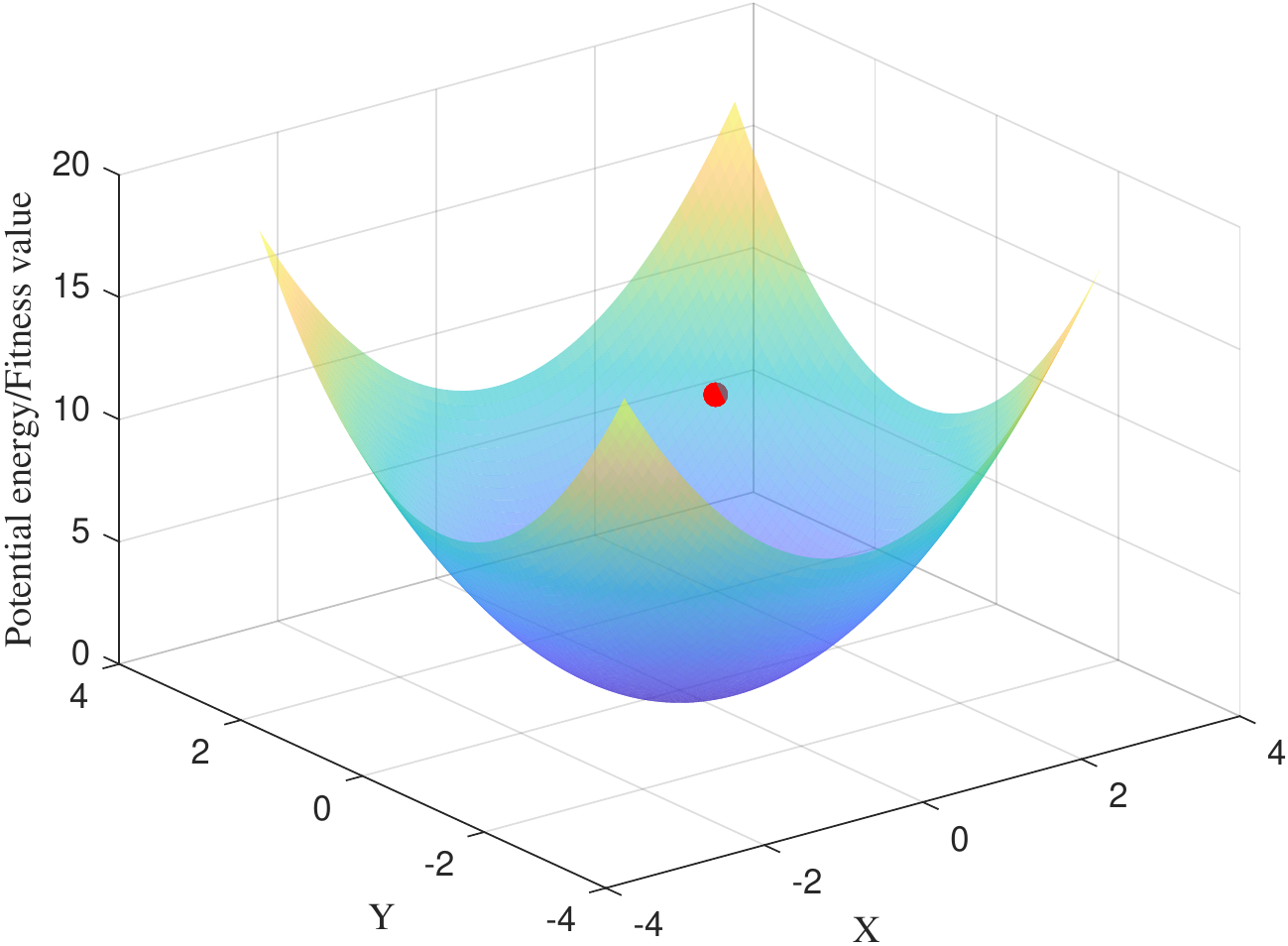} }
\subfigure[\tiny{Particle trajectories (without second-order estimation FE=200)}]{\includegraphics[width=4.5cm,angle=0]{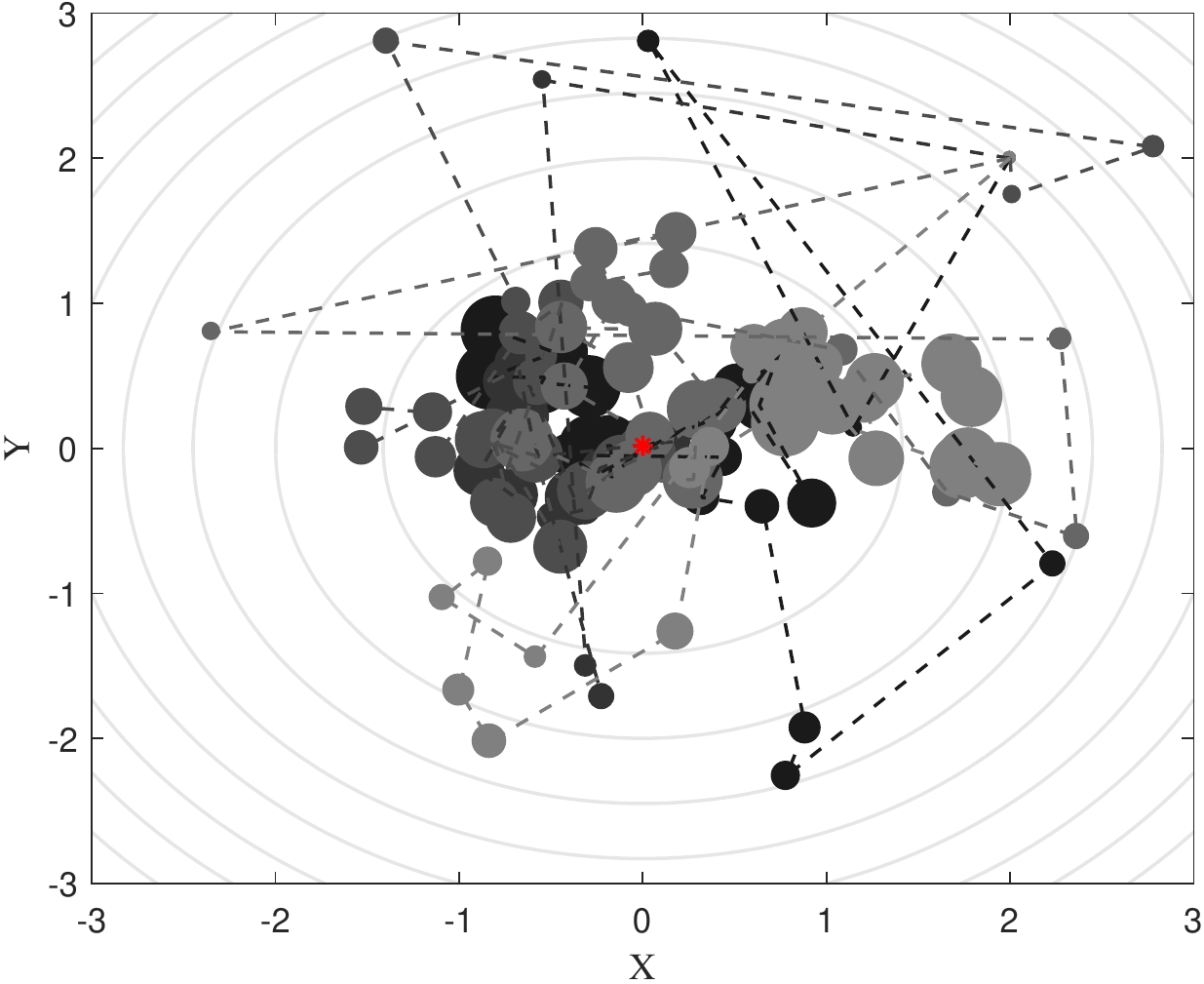} }
\subfigure[\tiny{Modulus of wave function $|\psi(x)|^{2}$ (FE=200)}]{\includegraphics[width=4.5cm,angle=0]{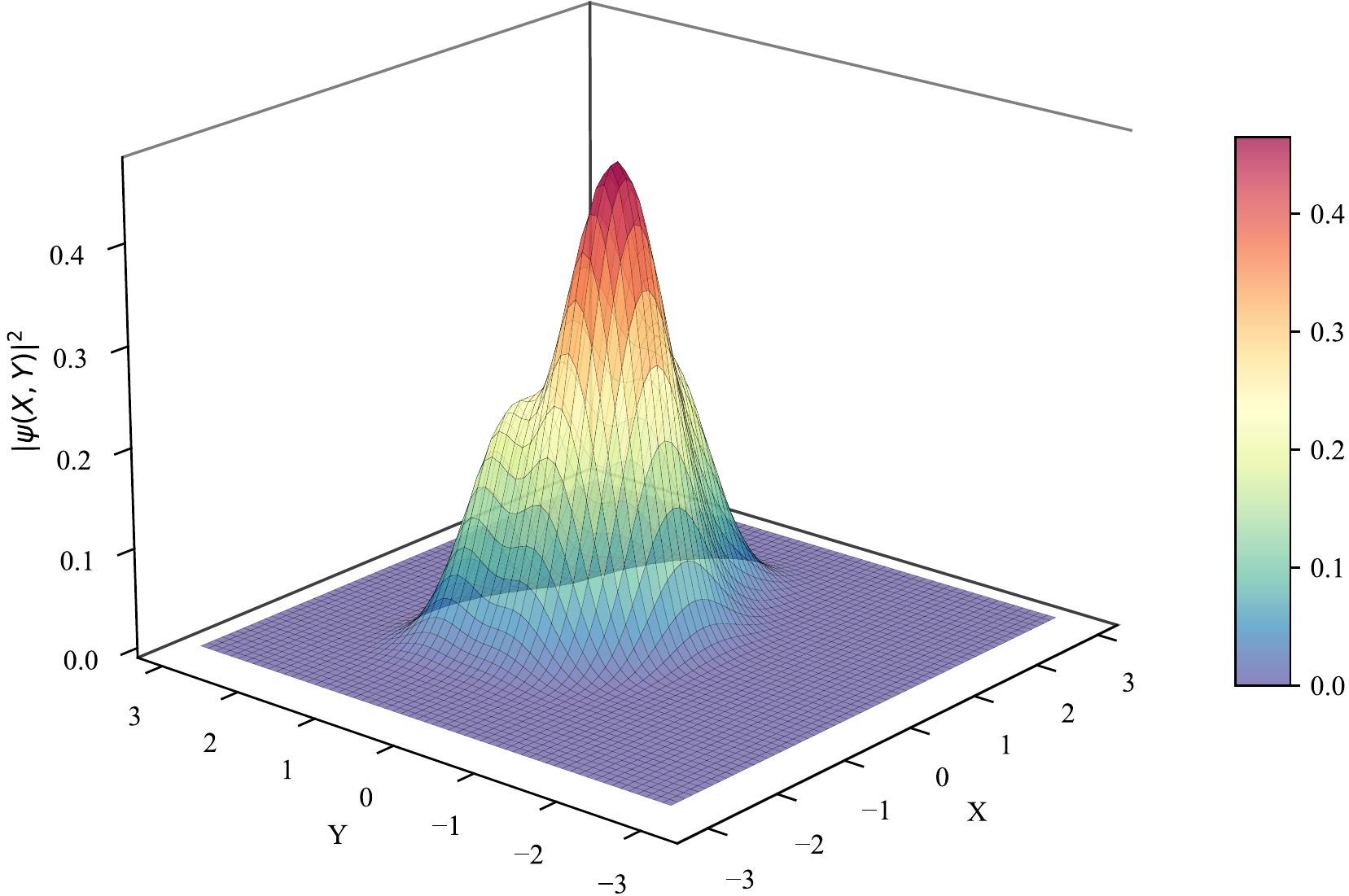} }

\subfigure[\tiny{Paraboloid function (FE=1)}]{\includegraphics[width=4.5cm,angle=0]{QHO_Initial.pdf} }
\subfigure[\tiny{Particle trajectories (with second-order estimation FE=200)}]{\includegraphics[width=4.5cm,angle=0]{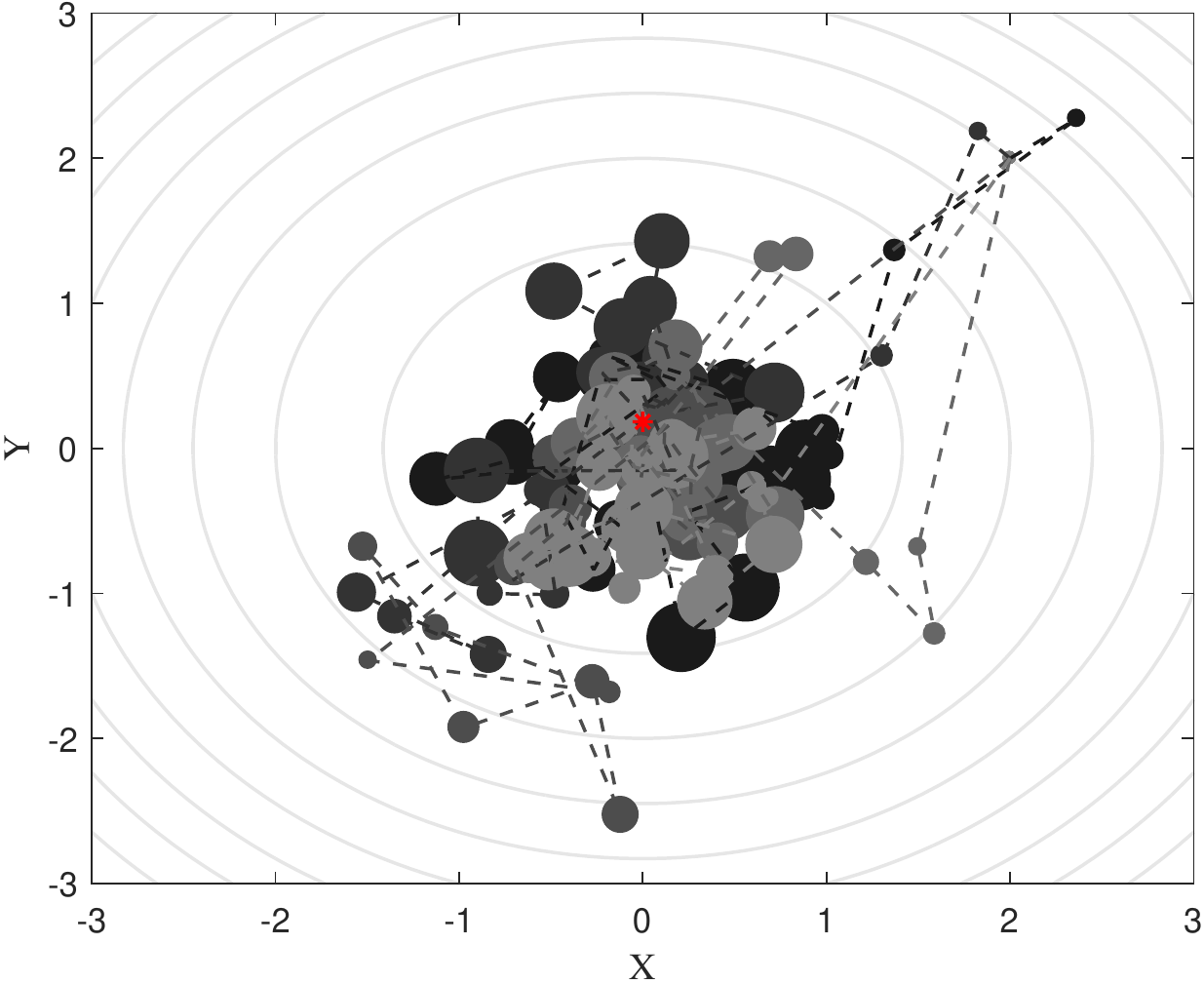} }
\subfigure[\tiny{Modulus of wave function $|\psi(x)|^{2}$ (FE=200)}]{\includegraphics[width=4.5cm,angle=0]{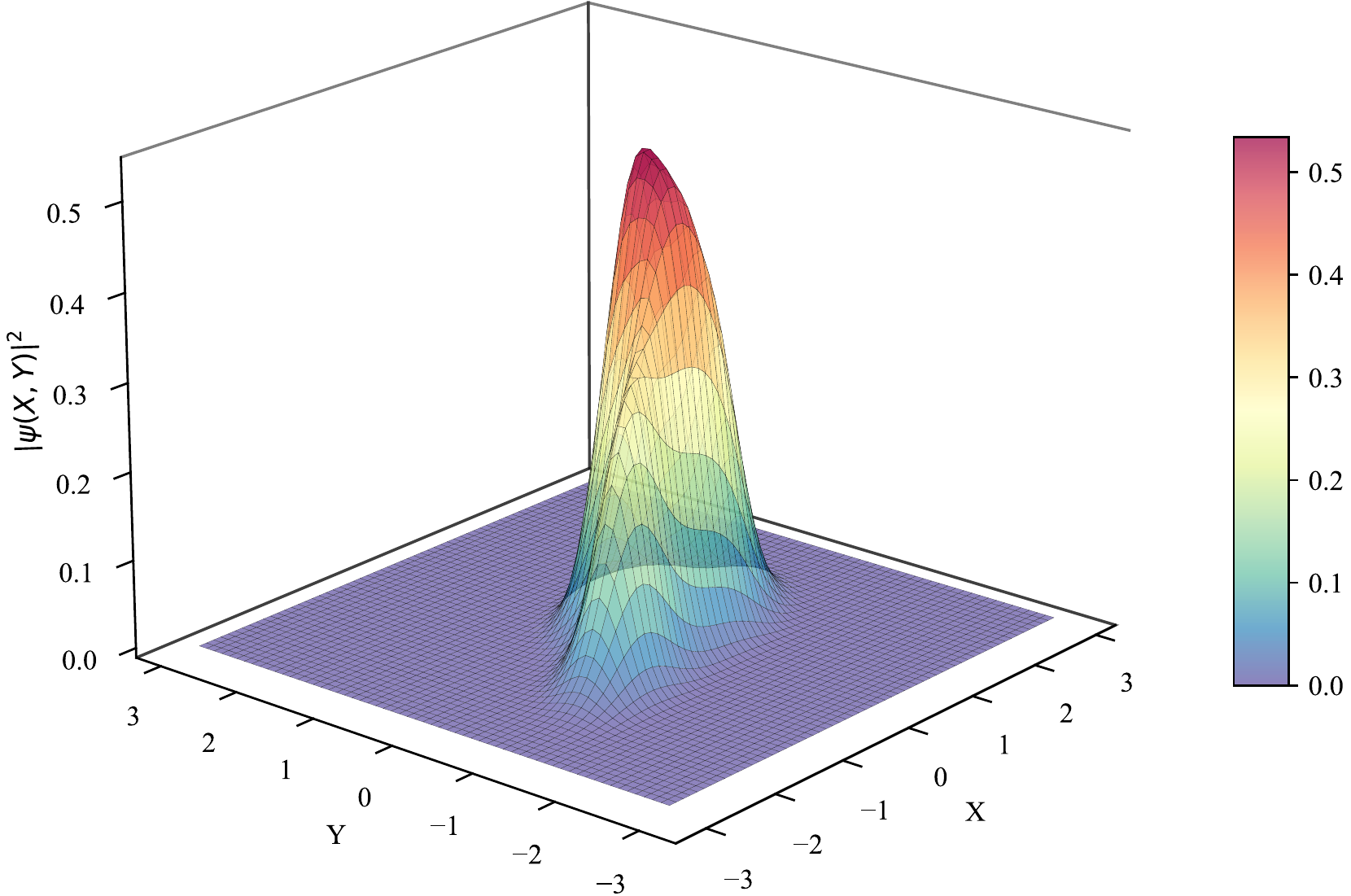} }
\caption{Evolution of algorithm under the different approximation of the paraboloid function}
\label{fig:QHOtrack}
\end{figure}

To investigate the evolutionary process of the BIP, the double-well function and paraboloid function are employed for benchmarking the modulus of the wave function $|\psi(x)|^{2}$ of BIP in the optimization. The dimension of the problem is set as 2; the size of the population is set as 5; The MaxFES is set as 200; the initialization position is (2,2); The positions of each copy of particles is recorded when it moves and the particle trajectories is plotted in the subgraphs (b) and (e) of Fig.~\ref{fig:DWtrack} and Fig.~\ref{fig:QHOtrack}, in which the initial radius of the particle is the smallest and increases with the movement of the particle. Moreover, the different particles are marked with different grayscales. According to the histogram of all particles the modulus of the wave function $|\psi(x)|^{2}$  is calculated and normalized in subgraphs (c) and (f) of Fig.~\ref{fig:DWtrack} and Fig.~\ref{fig:QHOtrack}.

It can be seen in Fig.~\ref{fig:DWtrack} (a) that the objective function is a simple double-well function with one global optima and three local optima. To demonstrate the dynamic behaviour based on the first- and second-order approximations, the initial position of particles is set at the local optimal position (nearby (2,2)) farthest from the global optimal (nearby (-2,-2)). The experimental result of BIP without second-order approximation is demonstrated in Fig.~\ref{fig:DWtrack} (b), in which the particles move in the search space only through the basic sampling behaviour and the quantum tunneling behaviour. After 200 function evaluations, only one particle passes through the largest barrier and enters the optimal solution region. In Fig.~\ref{fig:DWtrack} (c), the largest value of the modulus of the wave function is near (-2,2), which indicates that the local optimal has a large probability to become the solution to the optimization problem.

In contrast, as the experimental result of BIP with second-order approximation shown in Fig.~\ref{fig:DWtrack} (e), due to the mean position substitution behaviour, the distribution of particles is more concentrated, and the modulus of the wave function is concentrated in (-2,-2), which indicates that the current solutions is the most likely solution of the optimization problem.

To further illustrate the effect of mean substitution, a simple paraboloid function is selected as the test function. As shown in Fig.~\ref{fig:QHOtrack} (b) and (e), when there are second-order components in the objective optimization problem, the mean replacement is beneficial to the global optimization problem. In Fig.~\ref{fig:QHOtrack} (c), the modulus of the wave function of the BIP with the mean position replacement is more concentrated at the global optima than that in Fig.~\ref{fig:QHOtrack} (f), which means that the probability of obtaining the optimal solution is higher.

In general, BIP obtains diverse solutions through quantum tunneling and obtains a better global search ability through mean substitution.

\subsection{Investigation of transmission probabilities}

\begin{figure}[H]
\centering
\subfigure[\tiny{F1}]{\includegraphics[width=3.5cm,angle=0]{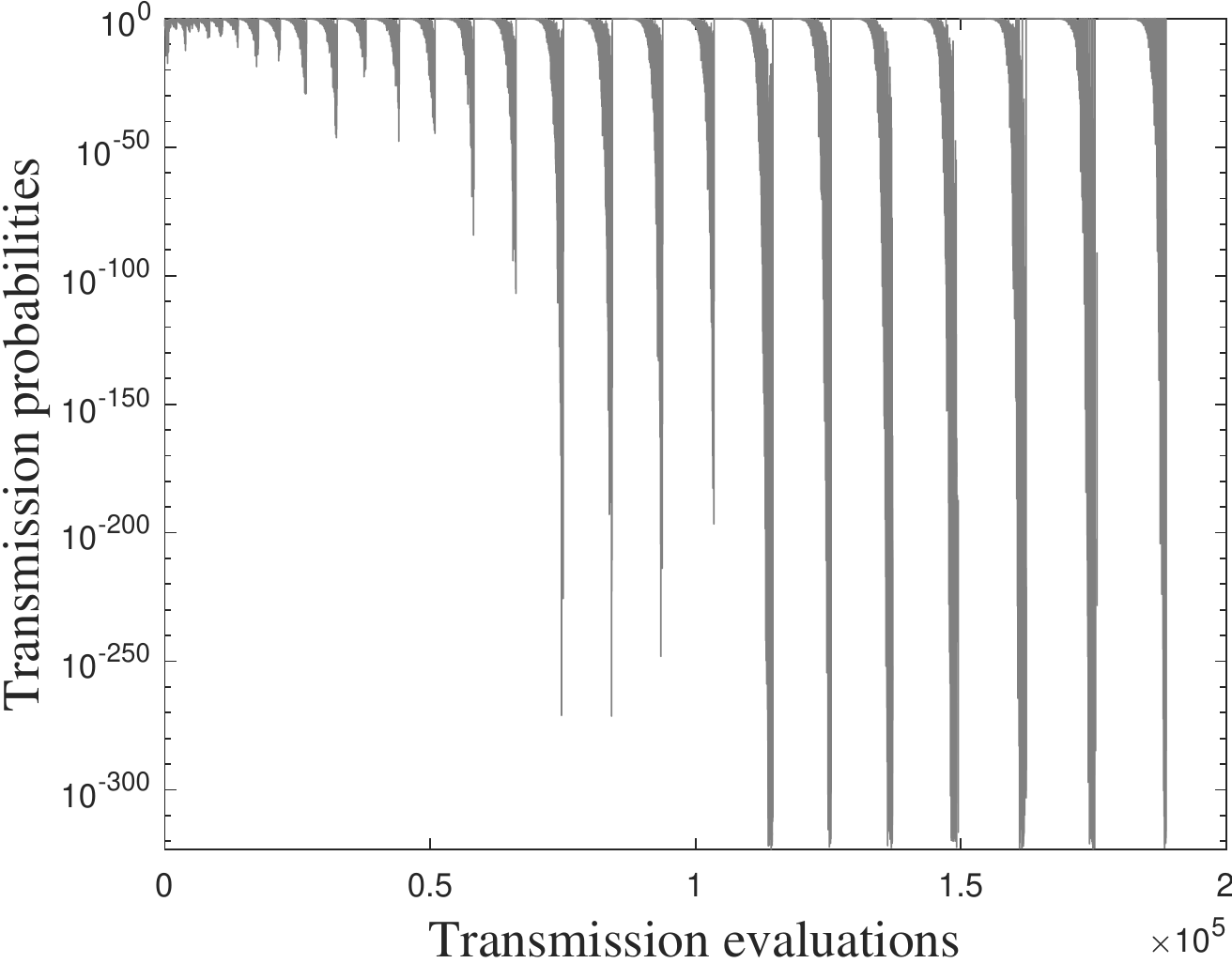} }
\subfigure[\tiny{F2}]{\includegraphics[width=3.5cm,angle=0]{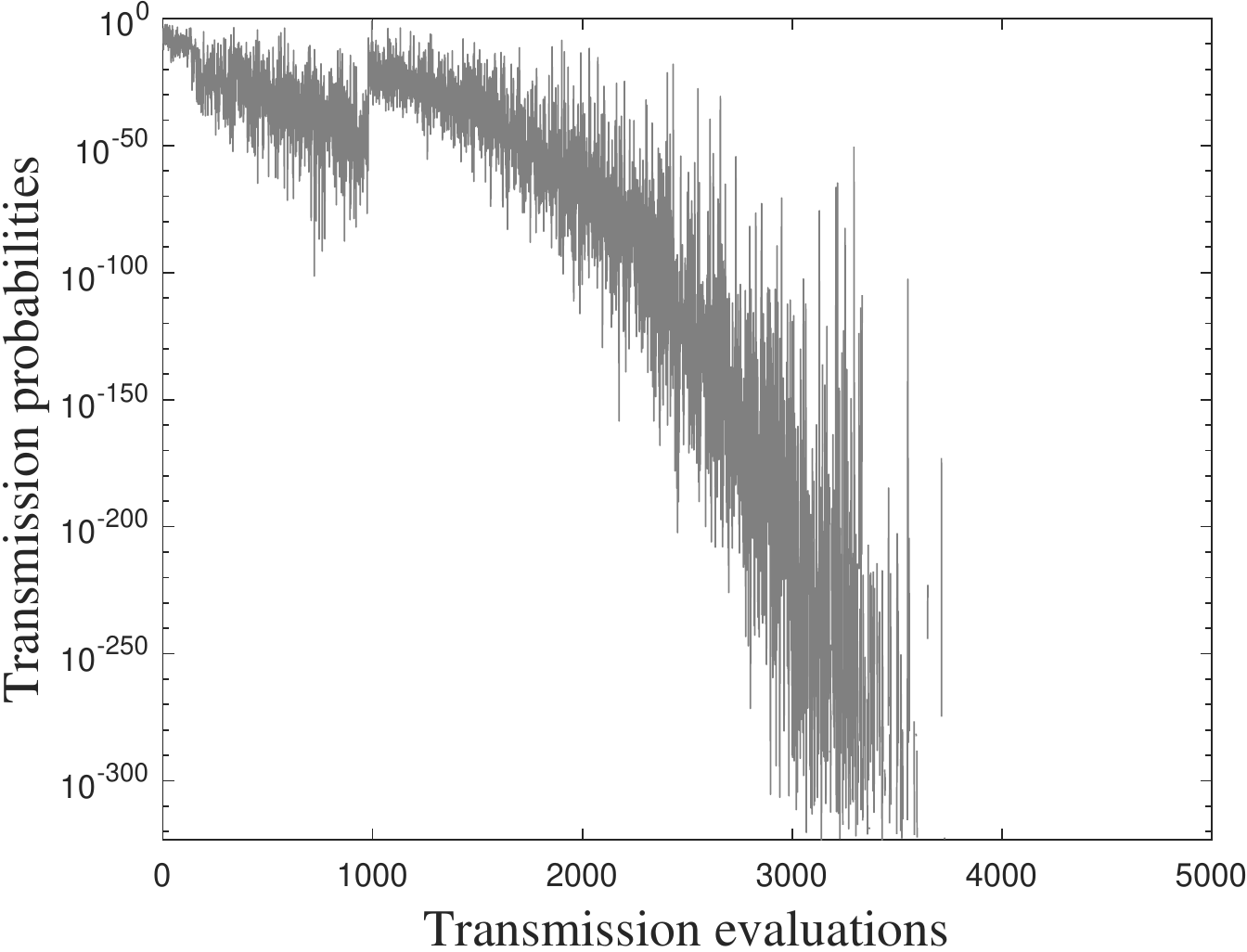} }
\subfigure[\tiny{F3}]{\includegraphics[width=3.5cm,angle=0]{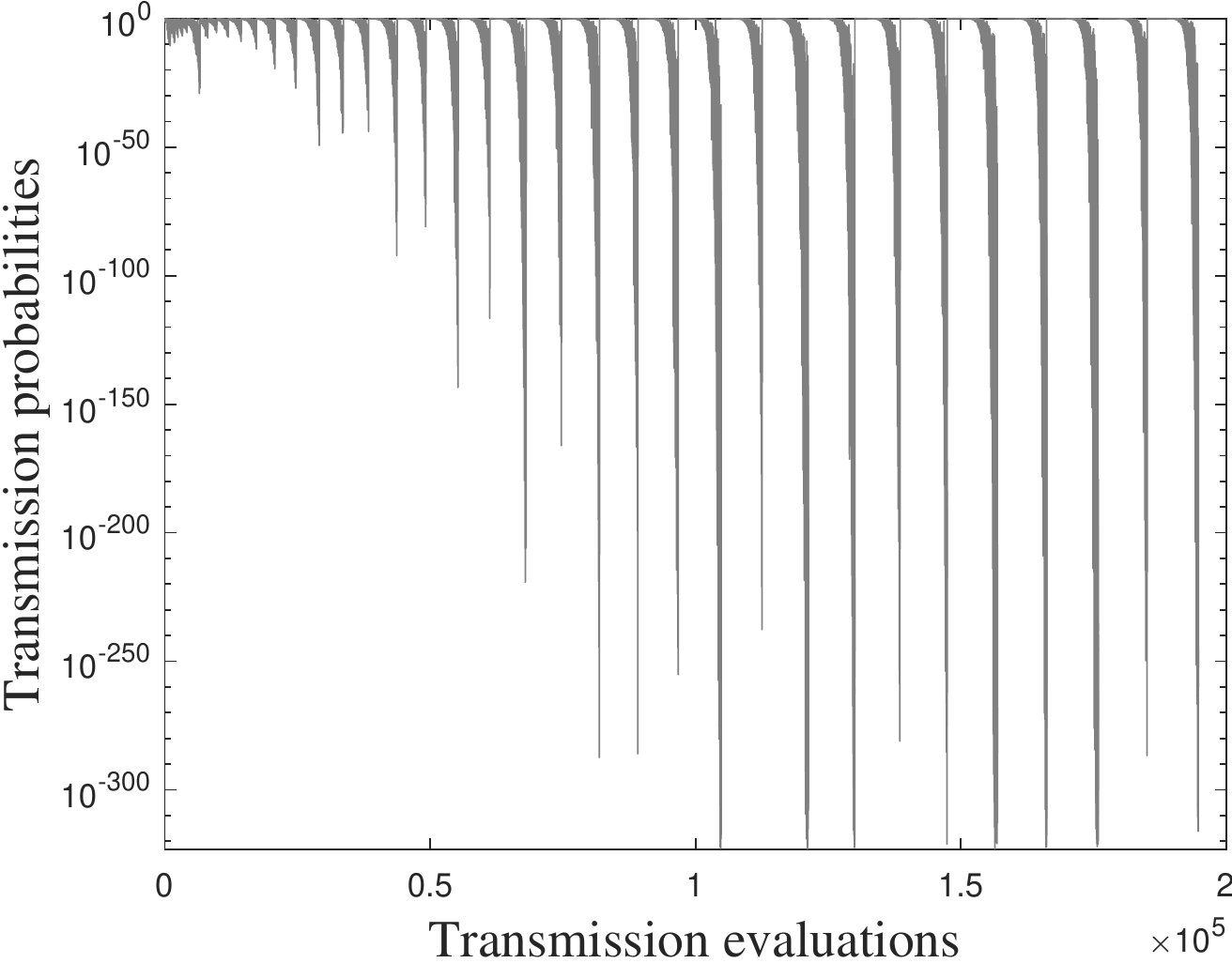}}
\subfigure[\tiny{F4}]{\includegraphics[width=3.5cm,angle=0]{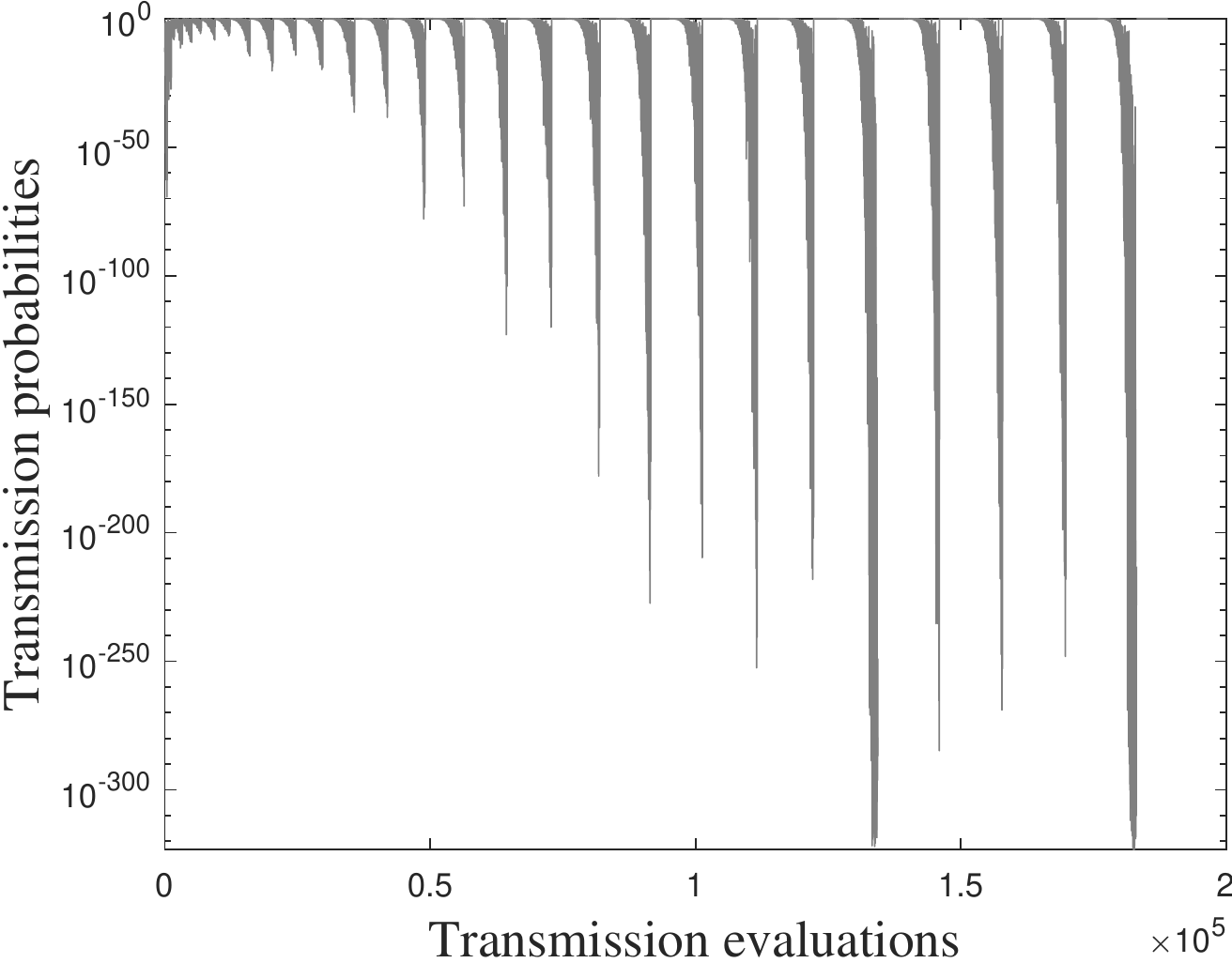} }

\subfigure[\tiny{F5}]{\includegraphics[width=3.5cm,angle=0]{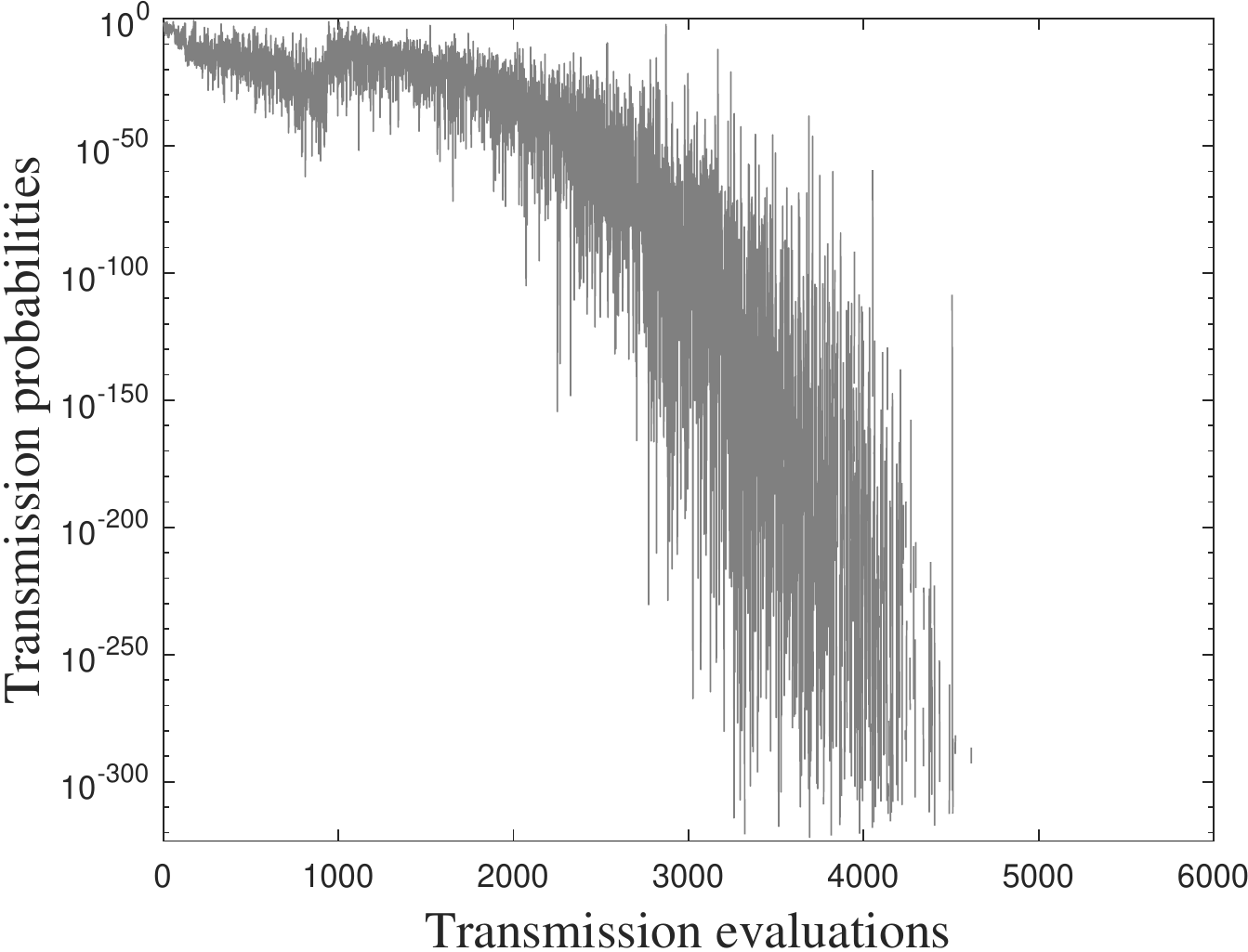} }
\subfigure[\tiny{F6}]{\includegraphics[width=3.5cm,angle=0]{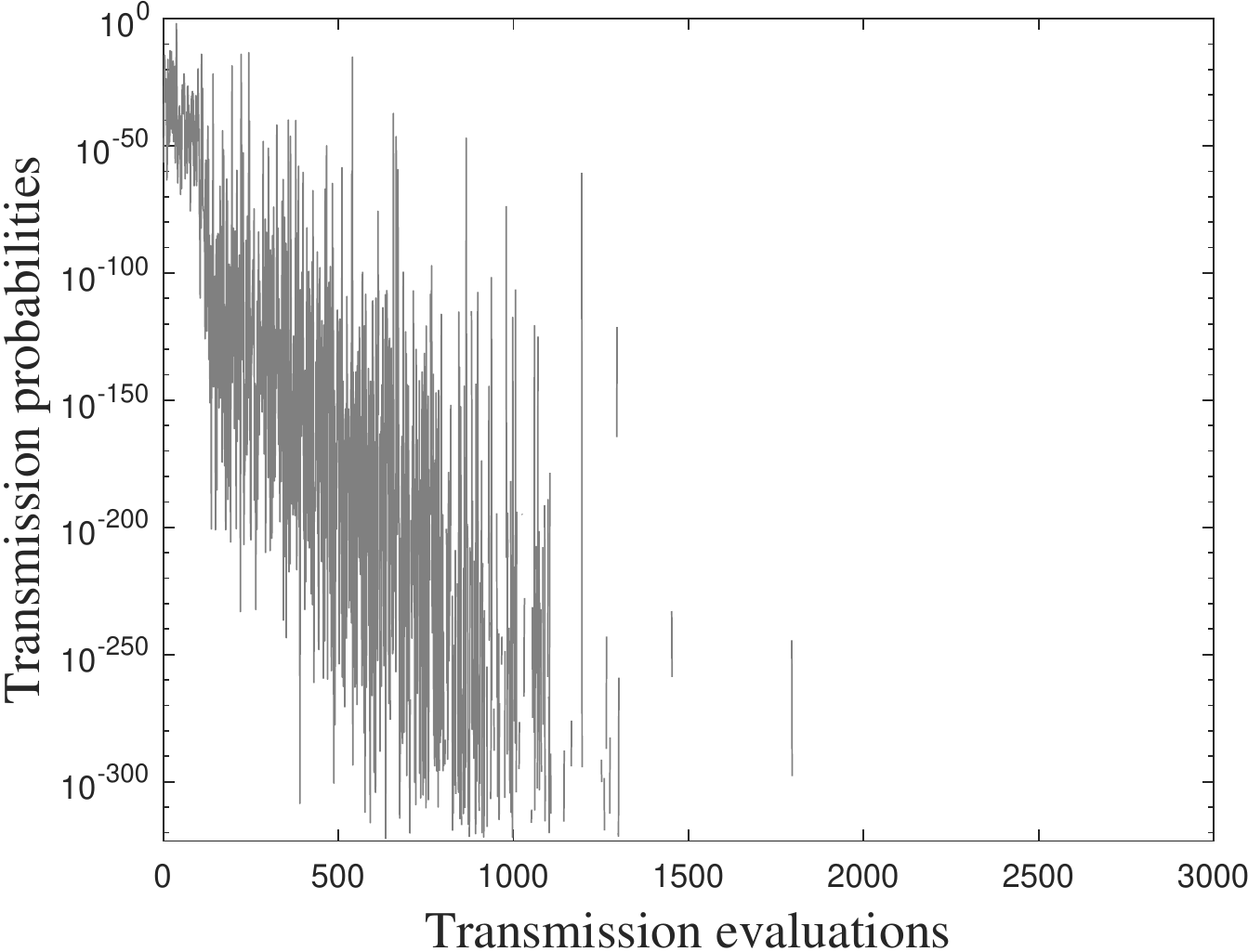} }
\subfigure[\tiny{F7}]{\includegraphics[width=3.5cm,angle=0]{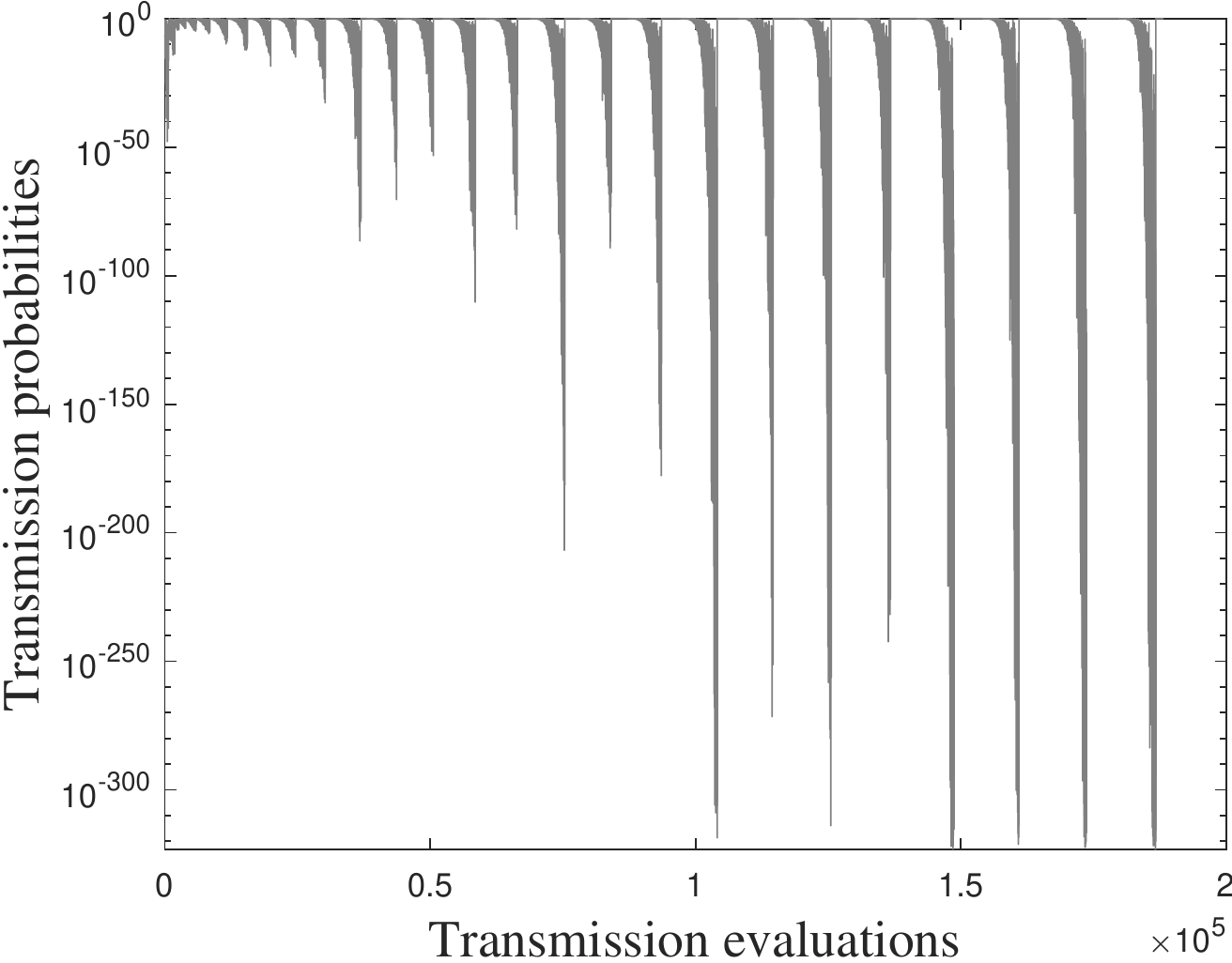} }
\subfigure[\tiny{F8}]{\includegraphics[width=3.5cm,angle=0]{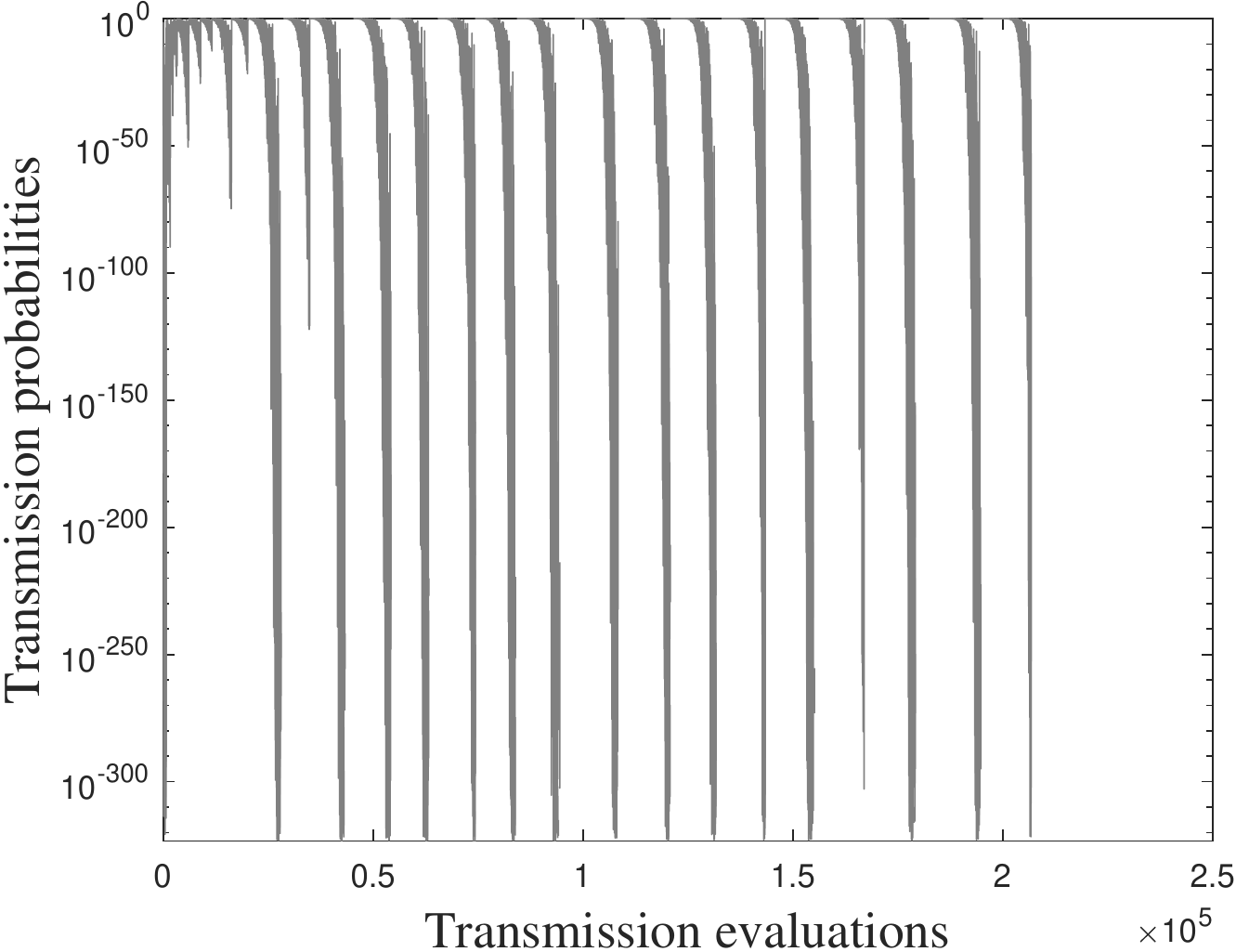} }

\subfigure[\tiny{F9}]{\includegraphics[width=3.5cm,angle=0]{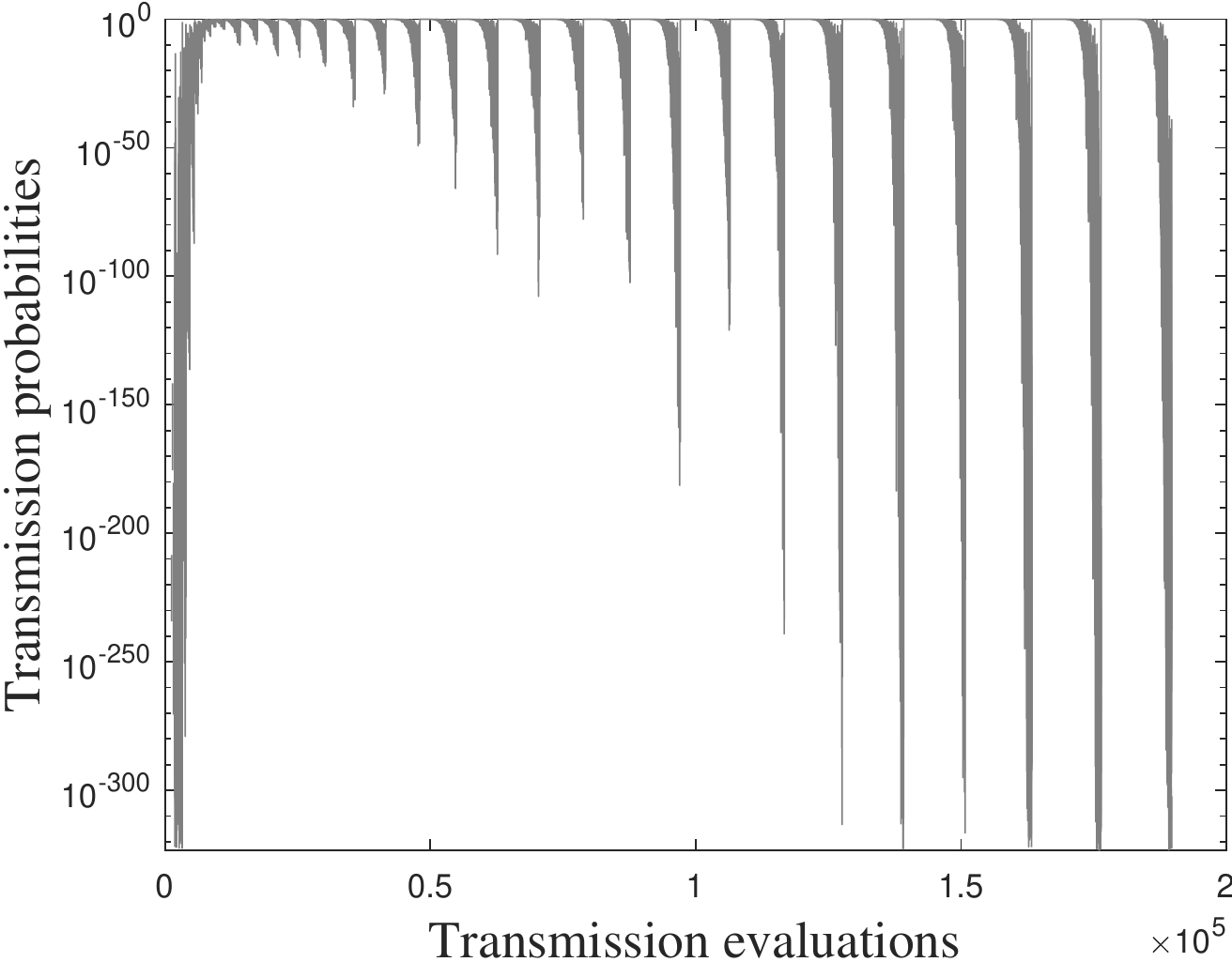} }
\subfigure[\tiny{F10}]{\includegraphics[width=3.5cm,angle=0]{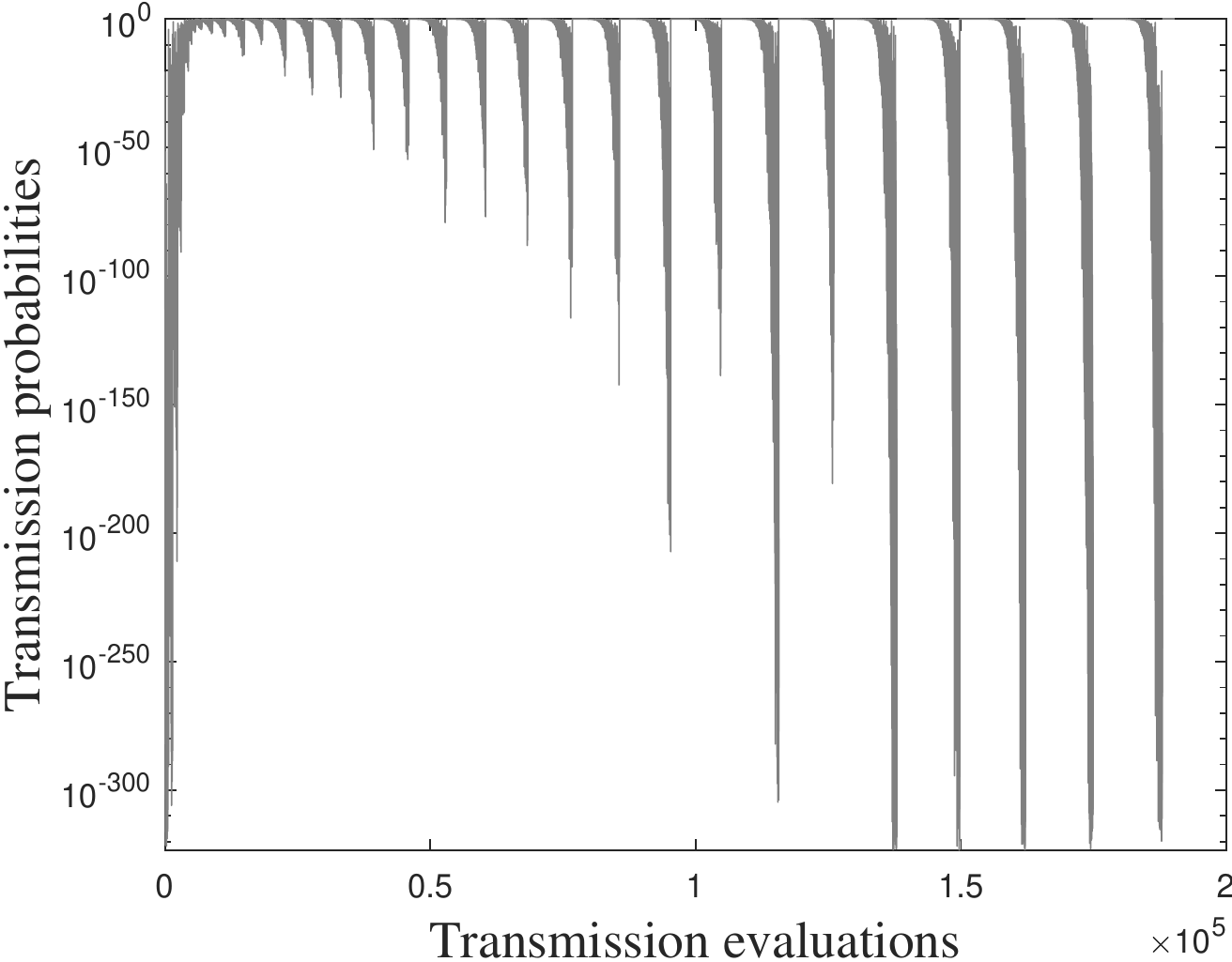} }
\subfigure[\tiny{F11}]{\includegraphics[width=3.5cm,angle=0]{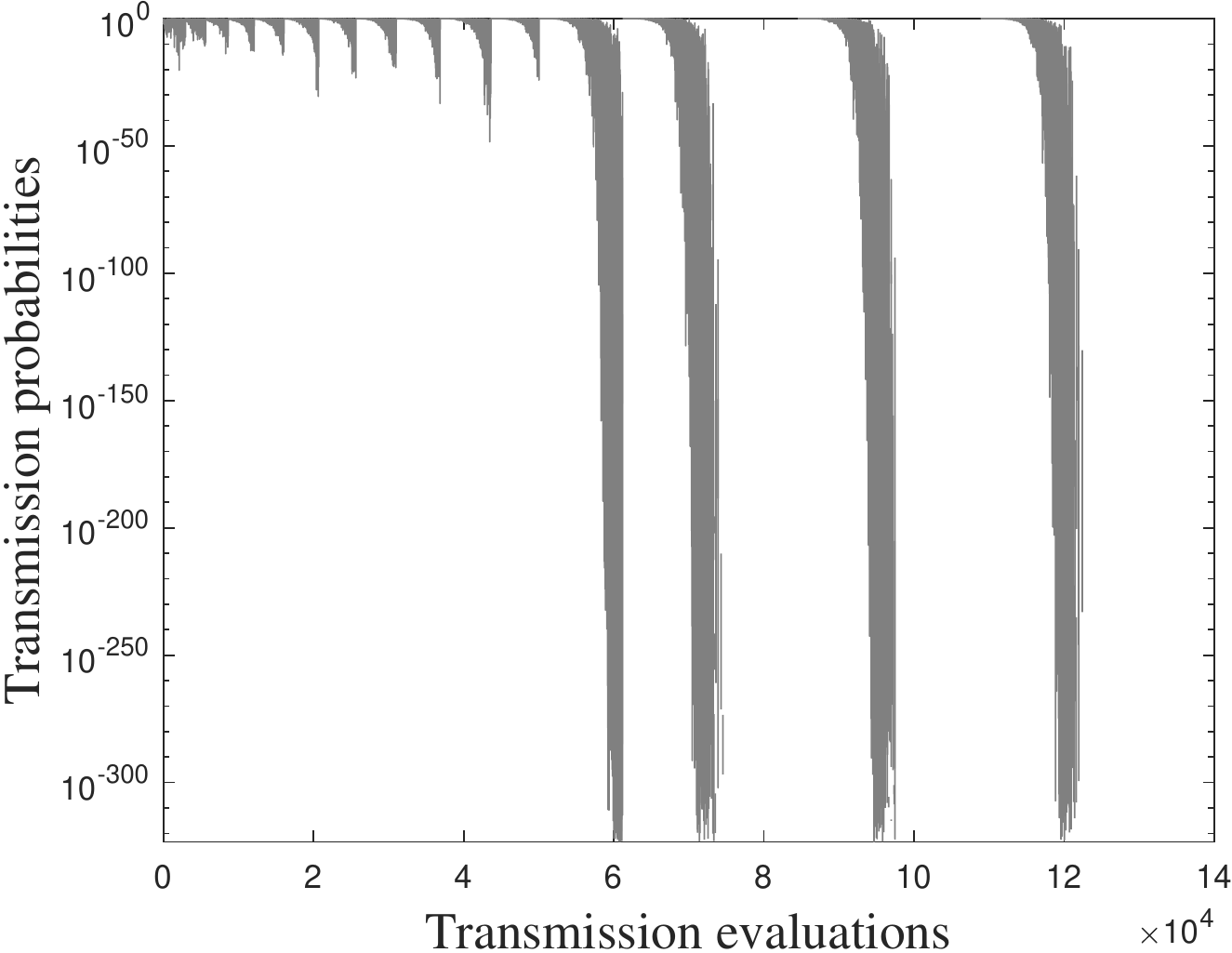} }
\subfigure[\tiny{F12}]{\includegraphics[width=3.5cm,angle=0]{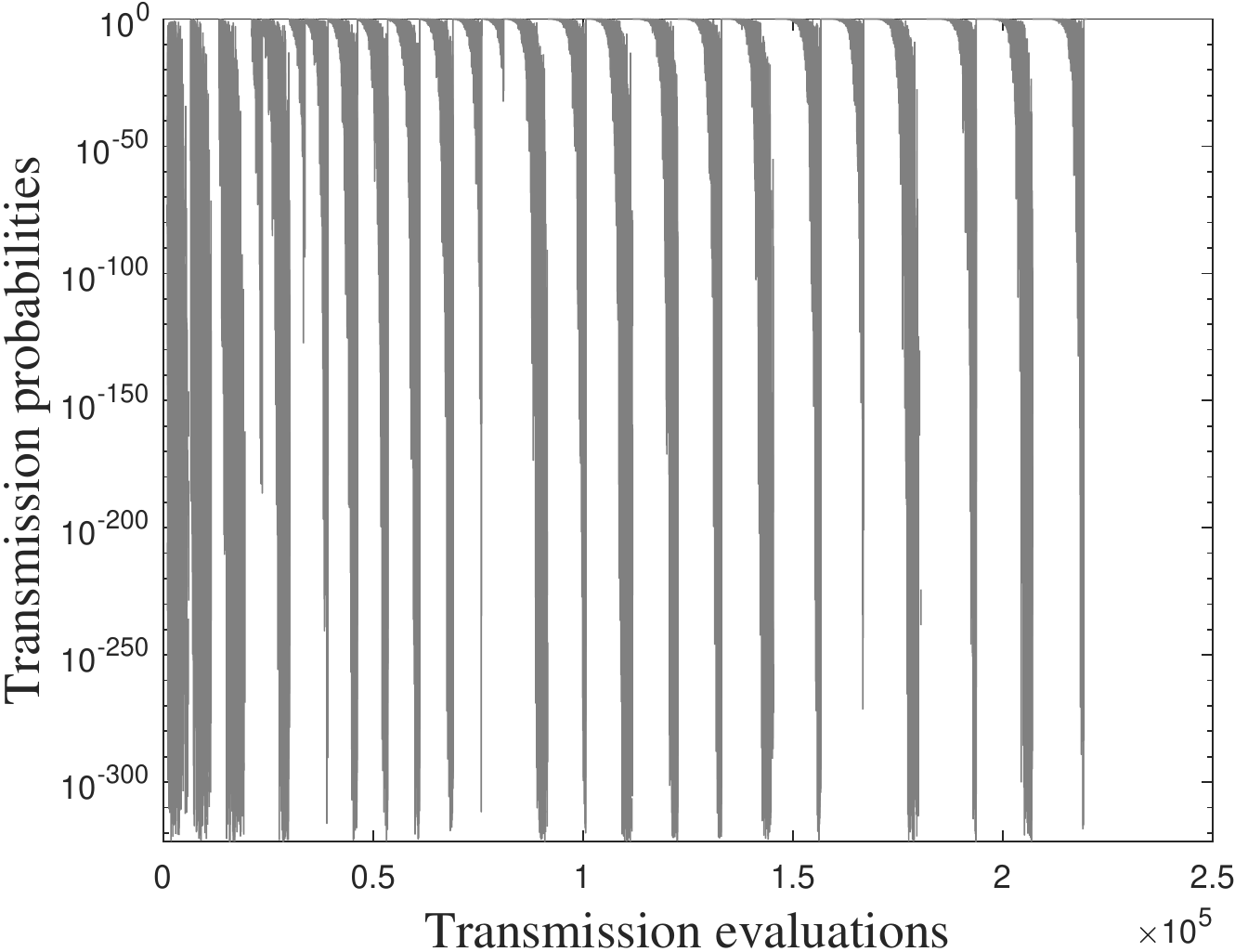} }
\caption{Transmission probabilities of quantum tunneling on 12 benchmark functions}
\label{fig:trans}
\end{figure}

In this subsection, the transmission probabilities of quantum tunneling in the BIP algorithm is studied on 12 benchmark functions. The transmission probabilities for the 30D problems are demonstrated in Fig.~\ref{fig:trans}.

As shown in Fig.~\ref{fig:trans}, the cyclic decrease of transmission probabilities can be clearly observed during the optimization process. This is because the particle energy $\Gamma$ is reduced exponentially according to Eq.~(\ref{equ:annealingC}) in the iterative process of each scale. It can be found that BIP successfully found the global optima in experiment of next subsection in Tables~\ref{tab:ME100D} except F2, F5 and F6. Correspondingly, in Fig.~\ref{fig:trans}, when the solution can be successfully obtained, the algorithm will undergo a multi-scale annealing process.

However, it is worth noting that the rate of decrease of the transmission probabilities in each scale is very fast, which means that quantum tunneling is ineffective most of the time during the search of this scale. This shows that Eq.~(\ref{equ:annealingC}) needs to be adjusted to adapt to the drastic changes in the fitness value in the numerical optimization. Since the purpose of this work is to study the physical interpretation of the basic search mechanism, the discussion is not continued here.

\subsection{Comparison with other schemes}
\label{otherscheme}
In this subsection, BIP is compared with 4 other schemes including the bare-bones particle swarm optimization (BBPSO) \cite{kennedy2003bare}, the bare-bones fireworks algorithm (BBFWA) \cite{li2018bare} and Gaussian bare-bones differential evolution (GBDE) \cite{wang2013gaussian}. All experiments with these schemes were conducted under the same conditions as those for all the schemes. The parameter settings used for BIP, BBPSO, BBFWA and GBDE are given in Table~\ref{tab:parameters}.

\subsubsection{Mean errors}

The best result, mean errors,standard deviations and success rete (Sr) for the 10D, 30D, and 50D problems are listed in Tables~\ref{tab:ME30D}, ~\ref{tab:ME60D}, and ~\ref{tab:ME100D}, respectively, where the minimum values for each function are highlighted. The rankings of the mean errors and standard deviations of the seven algorithms were calculated separately for the unimodal and multimodal functions.

\begin{table}[H]
  \centering
\scriptsize
  \setlength{\tabcolsep}{3.5pt}
  \caption{Comparison of mean errors and standard deviations of BIP, BBPSO, BBFWA and GBDE under 30D benchmark functions. The stopping condition for all schemes is set at MaxFES= 300000. The experiments are repeated 51 times individually.}
     \begin{tabular}{ccccccccc}
    \toprule
    Algorithm & AR. F1-F6 & Item  & F1    & F2    & F3    & F4    & F5    & F6 \\
    \midrule
    \multirow{4}[2]{*}{BIP} & \multirow{4}[2]{*}{3.17 } & Best  & 0.00E+00 & 8.95E+00 & \textbf{4.44E-15} & 1.26E-31 & 3.53E+00 & 6.88E+03 \\
          &       & Mean  & 2.66E-03 & 1.50E+02 & 4.92E-01 & 3.78E-01 & 2.31E+01 & 7.78E+03 \\
          &       & Std   & 5.14E-03 & 4.95E+01 & 7.84E-01 & 5.21E-01 & 8.38E+00 & 2.66E+02 \\
          &       & Sr    & 76.47\% & 0.00\% & 68.63\% & 23.53\% & 0.00\% & 0.00\% \\
    \midrule
    \multirow{4}[2]{*}{BBPSO} & \multirow{4}[2]{*}{2.50 } & Best  & 0.00E+00 & 2.39E+01 & 7.99E-15 & 2.80E-30 & 3.39E-14 & 1.78E+03 \\
          &       & Mean  & 1.14E-02 & 5.23E+01 & 8.85E-01 & 3.30E+00 & \textbf{9.24E-13} & 3.05E+03 \\
          &       & Std   & 1.52E-02 & 1.17E+01 & 9.88E-01 & 2.38E+00 & \textbf{2.34E-12} & 5.97E+02 \\
          &       & Sr    & 37.25\% & 0.00\% & 50.98\% & 3.92\% & \textbf{100.00\%} & 0.00\% \\
    \midrule
    \multirow{4}[2]{*}{BBFWA} & \multirow{4}[2]{*}{3.17 } & Best  & 1.07E-13 & 5.17E+01 & 2.46E-07 & 5.44E-01 & 1.57E-01 & 3.14E+03 \\
          &       & Mean  & 1.18E-02 & 9.97E+01 & 1.64E-01 & 1.04E+01 & 1.79E+00 & 4.94E+03 \\
          &       & Std   & 1.03E-02 & 2.74E+01 & 5.34E-01 & 7.86E+00 & 1.43E+00 & 5.39E+02 \\
          &       & Sr    & 25.49\% & 0.00\% & 88.24\% & 0.00\% & 0.00\% & 0.00\% \\
    \midrule
    \multirow{4}[2]{*}{GBDE} & \multirow{4}[2]{*}{1.17 } & Best  & \textbf{0.00E+00} & \textbf{0.00E+00} & 7.99E-15 & \textbf{1.50E-32} & \textbf{9.57E-15} & \textbf{3.82E-04} \\
          &       & Mean  & \textbf{2.18E-18} & \textbf{2.69E+00} & \textbf{7.99E-15} & \textbf{1.50E-32} & 1.04E-06 & \textbf{2.02E+02} \\
          &       & Std   & \textbf{1.55E-17} & \textbf{1.32E+00} & \textbf{0.00E+00} & \textbf{1.38E-47} & 7.43E-06 & \textbf{1.45E+02} \\
          &       & Sr    & \textbf{100.00\%} & \textbf{3.92\%} & \textbf{100.00\%} & \textbf{100.00\%} & 98.04\% & 0.00\% \\
    \midrule
    Algorithm & AR. F7-F12 & Item  & F7    & F8    & F9    & F10   & F11   & F12 \\
    \midrule
    \multirow{4}[2]{*}{BIP} & \multirow{4}[2]{*}{2.17 } & Best  & 3.53E-163 & 6.45E-25 & 3.65E-160 & \textbf{0.00E+00} & 5.79E-10 & \textbf{4.33E-37} \\
          &       & Mean  & 6.28E-161 & 1.49E-21 & 2.75E-158 & \textbf{4.95E-31} & 2.75E-09 & \textbf{1.06E-34} \\
          &       & Std   & 1.16E-160 & 2.48E-21 & 4.32E-158 & \textbf{3.53E-30} & 1.72E-09 & \textbf{3.95E-34} \\
          &       & Sr    & 100.00\% & 100.00\% & 100.00\% & \textbf{100.00\%} & 100.00\% & \textbf{100.00\%} \\
    \midrule
    \multirow{4}[2]{*}{BBPSO} & \multirow{4}[2]{*}{1.50 } & Best  & \textbf{1.22E-247} & \textbf{9.48E-245} & \textbf{2.45E-241} & 1.39E-28 & \textbf{0.00E+00} & 4.92E-26 \\
          &       & Mean  & \textbf{2.63E-236} & \textbf{8.03E-236} & \textbf{5.41E-231} & 2.11E-26 & \textbf{0.00E+00} & 4.26E-21 \\
          &       & Std   & \textbf{0.00E+00} & \textbf{0.00E+00} & \textbf{0.00E+00} & 1.24E-25 & \textbf{0.00E+00} & 1.31E-20 \\
          &       & Sr    & \textbf{100.00\%} & \textbf{100.00\%} & \textbf{100.00\%} & 100.00\% & \textbf{100.00\%} & 100.00\% \\
    \midrule
    \multirow{4}[2]{*}{BBFWA} & \multirow{4}[2]{*}{3.67 } & Best  & 2.45E-15 & 3.06E-12 & 2.27E-10 & 6.46E-13 & 3.24E-10 & 1.22E-13 \\
          &       & Mean  & 2.26E-13 & 1.13E-09 & 6.71E-09 & 6.82E-11 & 1.54E-09 & 6.91E-12 \\
          &       & Std   & 4.66E-13 & 1.96E-09 & 9.40E-09 & 1.32E-10 & 5.14E-10 & 9.85E-12 \\
          &       & Sr    & 100.00\% & 100.00\% & 100.00\% & 100.00\% & 100.00\% & 100.00\% \\
    \midrule
    \multirow{4}[2]{*}{GBDE} & \multirow{4}[2]{*}{2.67 } & Best  & 3.27E-59 & 1.55E-58 & 1.88E-54 & 0.00E+00 & 5.29E-142 & 8.09E-03 \\
          &       & Mean  & 3.48E-57 & 2.63E-55 & 1.52E-51 & 1.48E-30 & 3.81E-127 & 5.86E-02 \\
          &       & Std   & 9.21E-57 & 7.83E-55 & 3.59E-51 & 7.42E-30 & 1.68E-126 & 6.00E-02 \\
          &       & Sr    & 100.00\% & 100.00\% & 100.00\% & 100.00\% & 100.00\% & 0.00\% \\
    \bottomrule
    \end{tabular}%
  \label{tab:ME30D}%
\end{table}%

\begin{table}[H]
  \centering
\scriptsize
  \setlength{\tabcolsep}{3.5pt}
  \caption{Comparison of mean errors and standard deviations of BIP, BBPSO, BBFWA and GBDE under 60D benchmark functions. The stopping condition for all schemes is set at MaxFES= 600000. The experiments are repeated 51 times individually.}
    \begin{tabular}{ccccccccc}
    \toprule
    Algorithm & AR. F1-F6 & Item  & F1    & F2    & F3    & F4    & F5    & F6 \\
    \midrule
    \multirow{4}[2]{*}{BIP} & \multirow{4}[2]{*}{3.33 } & Best  & 0.00E+00 & 1.89E+01 & 1.87E-14 & 1.79E-01 & 1.36E+01 & 1.76E+04 \\
          &       & Mean  & 2.17E-03 & 3.24E+02 & 2.27E+00 & 1.65E+00 & 5.26E+01 & 1.83E+04 \\
          &       & Std   & 4.14E-03 & 1.66E+02 & 7.62E-01 & 1.35E+00 & 2.98E+01 & \textbf{2.86E+02} \\
          &       & Sr    & 76.47\% & 0.00\% & 3.92\% & 0.00\% & 0.00\% & 0.00\% \\
    \midrule
    \multirow{4}[2]{*}{BBPSO} & \multirow{4}[2]{*}{2.67 } & Best  & 0.00E+00 & 8.95E+01 & 2.93E-14 & 3.63E+00 & 8.79E-13 & 5.25E+03 \\
          &       & Mean  & 1.05E-02 & 1.41E+02 & 2.05E+00 & 1.49E+01 & 3.65E-04 & 7.30E+03 \\
          &       & Std   & 1.77E-02 & 2.44E+01 & 1.33E+00 & 5.69E+00 & 2.55E-03 & 8.65E+02 \\
          &       & Sr    & 50.98\% & 0.00\% & 17.65\% & 0.00\% & 96.08\% & 0.00\% \\
    \midrule
    \multirow{4}[2]{*}{BBFWA} & \multirow{4}[2]{*}{3.00 } & Best  & 1.89E-15 & 1.55E+02 & 1.42E-08 & 1.18E+01 & 1.89E+00 & 9.30E+03 \\
          &       & Mean  & 6.23E-03 & 2.24E+02 & 8.67E-01 & 3.93E+01 & 6.42E+00 & 1.09E+04 \\
          &       & Std   & 6.94E-03 & 3.25E+01 & 8.73E-01 & 1.15E+01 & 2.80E+00 & 8.91E+02 \\
          &       & Sr    & 45.10\% & 0.00\% & 47.06\% & 0.00\% & 0.00\% & 0.00\% \\
    \midrule
    \multirow{4}[2]{*}{GBDE} & \multirow{4}[2]{*}{1.00 } & Best  & \textbf{0.00E+00} & \textbf{1.59E+01} & \textbf{1.51E-14} & \textbf{1.50E-32} & \textbf{3.04E-14} & \textbf{1.05E+03} \\
          &       & Mean  & \textbf{4.83E-04} & \textbf{2.71E+01} & \textbf{1.82E-14} & \textbf{1.78E-02} & \textbf{2.34E-11} & \textbf{1.88E+03} \\
          &       & Std   & \textbf{1.97E-03} & \textbf{5.75E+00} & \textbf{3.33E-15} & \textbf{8.91E-02} & \textbf{9.50E-11} & 5.01E+02 \\
          &       & Sr    & \textbf{94.12\%} & 0.00\% & \textbf{100.00\%} & \textbf{96.08\%} & \textbf{100.00\%} & 0.00\% \\
    \midrule
    Algorithm & AR. F7-F12 & Item  & F7    & F8    & F9    & F10   & F11   & F12 \\
    \midrule
    \multirow{4}[2]{*}{BIP} & \multirow{4}[2]{*}{2.00 } & Best  & \textbf{4.29E-208} & 1.11E-13 & \textbf{1.35E-203} & 0.00E+00 & 9.80E-11 & \textbf{5.04E-16} \\
          &       & Mean  & \textbf{8.06E-206} & 1.60E-10 & \textbf{2.28E-201} & 5.52E-28 & 1.28E-09 & \textbf{1.37E-14} \\
          &       & Std   & \textbf{0.00E+00} & 2.38E-10 & \textbf{0.00E+00} & 1.25E-27 & 7.91E-10 & \textbf{1.62E-14} \\
          &       & Sr    & \textbf{100.00\%} & 100.00\% & \textbf{100.00\%} & 100.00\% & 100.00\% & \textbf{100.00\%} \\
    \midrule
    \multirow{4}[2]{*}{BBPSO} & \multirow{4}[2]{*}{2.00 } & Best  & 1.35E-158 & \textbf{6.13E-155} & 4.75E-154 & 1.08E-25 & \textbf{1.35E-145} & 8.63E-06 \\
          &       & Mean  & 3.90E-150 & \textbf{1.06E-146} & 9.11E-144 & 4.50E-19 & \textbf{2.26E-118} & 9.44E-04 \\
          &       & Std   & 1.41E-149 & \textbf{4.44E-146} & 6.46E-143 & 3.16E-18 & \textbf{1.59E-117} & 2.42E-03 \\
          &       & Sr    & 100.00\% & \textbf{100.00\%} & 100.00\% & 100.00\% & \textbf{100.00\%} & 1.96\% \\
    \midrule
    \multirow{4}[2]{*}{BBFWA} & \multirow{4}[2]{*}{3.50 } & Best  & 1.01E-17 & 8.78E-09 & 4.68E-12 & 1.87E-15 & 3.53E-10 & 1.01E-13 \\
          &       & Mean  & 3.08E-16 & 1.14E-04 & 9.14E-11 & 1.44E-13 & 8.06E-10 & 1.70E-12 \\
          &       & Std   & 4.49E-16 & 1.69E-04 & 1.16E-10 & 2.48E-13 & 2.42E-10 & 1.79E-12 \\
          &       & Sr    & 100.00\% & 31.37\% & 100.00\% & 100.00\% & 100.00\% & 100.00\% \\
    \midrule
    \multirow{4}[2]{*}{GBDE} & \multirow{4}[2]{*}{2.50 } & Best  & 1.94E-39 & 3.63E-37 & 7.79E-34 & \textbf{0.00E+00} & 6.84E-91 & 2.91E+01 \\
          &       & Mean  & 9.72E-37 & 1.04E-34 & 2.29E-31 & \textbf{2.65E-28} & 3.14E-75 & 6.04E+01 \\
          &       & Std   & 2.54E-36 & 2.46E-34 & 5.34E-31 & \textbf{4.11E-28} & 1.95E-74 & 1.77E+01 \\
          &       & Sr    & 100.00\% & 100.00\% & 100.00\% & \textbf{100.00\%} & 100.00\% & 0.00\% \\
    \bottomrule
    \end{tabular}%
   \label{tab:ME60D}%
\end{table}%

\begin{table}[H]
  \centering
\scriptsize
  \setlength{\tabcolsep}{3.5pt}
  \caption{Comparison of mean errors and standard deviations of BIP, BBPSO, BBFWA and GBDE under 100D benchmark functions. The stopping condition for all schemes is set at MaxFES= 1000000. The experiments are repeated 51 times individually.}
    \begin{tabular}{ccccccccc}
    \toprule
    Algorithm & AR. F1-F6 & Item  & F1    & F2    & F3    & F4    & F5    & F6 \\
    \midrule
    \multirow{4}[2]{*}{BIP} & \multirow{4}[2]{*}{3.33 } & Best  & 2.22E-16 & \textbf{3.38E+01} & 2.29E+00 & 2.07E+00 & 4.64E+01 & 3.20E+04 \\
          &       & Mean  & 2.75E-03 & 6.45E+02 & 4.03E+00 & 4.41E+00 & 1.61E+02 & 3.29E+04 \\
          &       & Std   & 5.65E-03 & 2.82E+02 & 1.15E+00 & 1.34E+00 & 4.44E+01 & 4.38E+02 \\
          &       & Sr    & 78.43\% & 0.00\% & 0.00\% & 0.00\% & 0.00\% & 0.00\% \\
    \midrule
    \multirow{4}[2]{*}{BBPSO} & \multirow{4}[2]{*}{2.67 } & Best  & 4.44E-16 & 2.04E+02 & 1.21E+00 & 2.00E+01 & 2.71E-11 & 1.12E+04 \\
          &       & Mean  & 3.42E-02 & 2.66E+02 & 3.00E+00 & 3.72E+01 & 5.91E-04 & 1.31E+04 \\
          &       & Std   & 5.28E-02 & 3.63E+01 & 9.54E-01 & 8.98E+00 & 3.75E-03 & 1.24E+03 \\
          &       & Sr    & 50.98\% & 0.00\% & 0.00\% & 0.00\% & 84.31\% & 0.00\% \\
    \midrule
    \multirow{4}[2]{*}{BBFWA} & \multirow{4}[2]{*}{3.00 } & Best  & 1.22E-15 & 3.05E+02 & 6.98E-09 & 4.86E+01 & 4.01E+00 & 1.67E+04 \\
          &       & Mean  & 2.75E-03 & 4.03E+02 & 1.49E+00 & 7.92E+01 & 1.37E+01 & 1.90E+04 \\
          &       & Std   & 4.48E-03 & 5.48E+01 & 7.70E-01 & 1.85E+01 & 5.00E+00 & 1.14E+03 \\
          &       & Sr    & 70.59\% & 0.00\% & 17.65\% & 0.00\% & 0.00\% & 0.00\% \\
    \midrule
    \multirow{4}[2]{*}{GBDE} & \multirow{4}[2]{*}{1.00 } & Best  & \textbf{0.00E+00} & 5.97E+01 & \textbf{8.93E-13} & \textbf{7.23E-23} & \textbf{1.17E-10} & \textbf{4.32E+03} \\
          &       & Mean  & \textbf{0.00E+00} & \textbf{9.24E+01} & \textbf{1.40E-11} & \textbf{4.08E-01} & \textbf{3.94E-08} & \textbf{5.61E+03} \\
          &       & Std   & \textbf{0.00E+00} & \textbf{1.68E+01} & \textbf{1.69E-11} & \textbf{4.51E-01} & \textbf{9.74E-08} & 6.71E+02 \\
          &       & Sr    & \textbf{100.00\%} & 0.00\% & \textbf{100.00\%} & \textbf{47.06\%} & \textbf{100.00\%} & 0.00\% \\
    \midrule
    Algorithm & AR. F7-F12 & Item  & F7    & F8    & F9    & F10   & F11   & F12 \\
    \midrule
    \multirow{4}[2]{*}{BIP} & \multirow{4}[2]{*}{2.00 } & Best  & \textbf{8.61E-226} & 2.48E-09 & \textbf{1.37E-220} & \textbf{0.00E+00} & 1.80E-10 & 8.84E-07 \\
          &       & Mean  & \textbf{6.68E-224} & 1.82E-06 & \textbf{3.44E-218} & \textbf{2.16E-26} & 7.83E-10 & 1.03E-05 \\
          &       & Std   & \textbf{0.00E+00} & 2.15E-06 & \textbf{0.00E+00} & \textbf{1.83E-26} & 3.62E-10 & 1.01E-05 \\
          &       & Sr    & \textbf{100.00\%} & \textbf{100.00\%} & \textbf{100.00\%} & \textbf{100.00\%} & 100.00\% & 66.66\% \\
    \midrule
    \multirow{4}[2]{*}{BBPSO} & \multirow{4}[2]{*}{2.33 } & Best  & 2.57E-109 & \textbf{1.72E-105} & 7.35E-102 & 7.04E-22 & \textbf{2.46E-63} & 2.80E+00 \\
          &       & Mean  & 7.37E-101 & \textbf{5.96E-99} & 1.30E-94 & 1.32E-10 & \textbf{4.74E-33} & 1.35E+01 \\
          &       & Std   & 4.85E-100 & \textbf{2.33E-98} & 6.30E-94 & 9.41E-10 & \textbf{3.38E-32} & 8.52E+00 \\
          &       & Sr    & 100.00\% & \textbf{100.00\%} & 100.00\% & 100.00\% & \textbf{100.00\%} & 0.00\% \\
    \midrule
    \multirow{4}[2]{*}{BBFWA} & \multirow{4}[2]{*}{3.16 } & Best  & 9.15E-19 & 6.10E-06 & 5.47E-13 & 7.60E-16 & 2.22E-10 & \textbf{1.93E-12} \\
          &       & Mean  & 2.52E-17 & 1.63E-02 & 2.01E-11 & 1.36E-14 & 4.31E-10 & \textbf{9.42E-12} \\
          &       & Std   & 6.36E-17 & 1.99E-02 & 2.49E-11 & 1.54E-14 & 9.29E-11 & \textbf{6.68E-12} \\
          &       & Sr    & 100.00\% & 1.96\% & 100.00\% & 100.00\% & 100.00\% & \textbf{100.00\%} \\
    \midrule
    \multirow{4}[2]{*}{GBDE} & \multirow{4}[2]{*}{2.50 } & Best  & 3.85E-25 & 6.25E-23 & 1.61E-19 & 7.67E-22 & 5.39E-62 & 2.22E+02 \\
          &       & Mean  & 4.90E-23 & 5.88E-21 & 2.53E-17 & 3.19E-19 & 4.58E-50 & 3.33E+02 \\
          &       & Std   & 1.56E-22 & 9.91E-21 & 3.71E-17 & 1.58E-18 & 2.00E-49 & 4.86E+01 \\
          &       & Sr    & 100.00\% & 100.00\% & 100.00\% & 100.00\% & 100.00\% & 0.00\% \\
    \bottomrule
    \end{tabular}%
  \label{tab:ME100D}%
\end{table}%

On the multimodal functions, GBDE displayed excellent performance, and other algorithm have not performed as well, among them, the performance of BBPSO algorithm is significantly better. BBFWA and BIP are far behind the first two algorithms, ranking third and fourth respectively.

On the unimodal functions, all four algorithms can almost find the position of the global optimal solution, and the differences are limited to the calculation accuracy that each algorithm can achieve under the same number of iterations. It must be mentioned that the benchmark function F12 has higher requirements on the algorithm as the dimensionality increases. GBDE and BBPSO failed to find the optima on F12, BIP has $66.66\%$ opportunity to archive the best result on F12. According to the ranking evaluation, BIP ranked first and BBPSO exhibited stable performance and ranked second on the unimodal functions in all dimensions.

In general, as shown in the Tables~\ref{tab:ME30D}, ~\ref{tab:ME60D}, and ~\ref{tab:ME100D}, GBDE achieved the minimum error for 5 times of 30D problems, 7 times of 60D problems and 5 times of 100D problems (almost of those on multimodal functions), BIP achieved the minimum error for 3 times of the 30D problems, 3 times of 60D problems and 4 times of 100D problems which respectively (all of those on unimodal functions), whereas BBPSO achieved 4 times of the 30D problems, 2 times of 60D problems and 2 times of 100D problems. BBFWA achieved only 1 time on 100D problem. It shows that GBDE has good performance on multimodal functions while BIP has good performance on unimodoal functions. The BIP with the basic operators obtained similar performance in comparison with the other two bare-bone schemes (BBPSO and BBFWA).

\subsubsection{Investigation of the influence of problem dimensionality}

\begin{figure}[H]
\centerline{\includegraphics[width=15cm, angle=0]{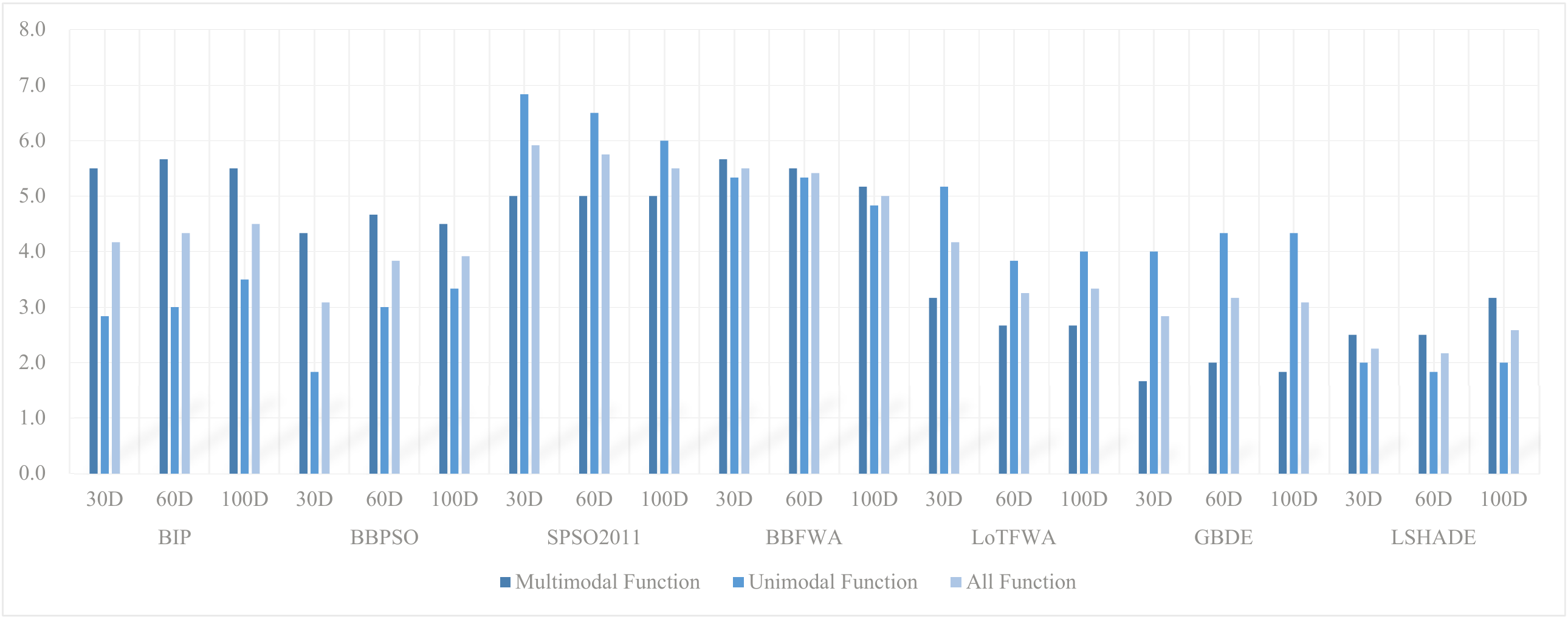} }
\caption{Comparison of ARs of BIP, BBPSO, BBFWA and GBDE on 30D, 60D and 100D benchmark functions.}
\label{fig:dimenstionRank}
\end{figure}

In this subsection, we report the results of our investigation on the performance of the algorithms on problems with increasing dimensionality. As a direct comparison of the mean errors in different dimensions would be unfair, we list the average rankings for the 30D to 100D problems in Fig.~\ref{fig:dimenstionRank}.

As demonstrated in the figure, BBFWA exhibited a positive ranking encasement both for the multiimodal and unimodal functions when the problem dimensionality increased from 30D to 100D. In contrast, the rankings of BIP, BBPSO and GBDE worsened as the dimensionality increased.

In general, except for BBFWA, the results for BIP and two other bare-bone schemes show weak adaptability to increasing dimensionality.

\subsection{Discussion}
\label{Discussion}

In this section, we summarize the aforementioned experimental results and focus on analyzing the different search mechanisms of the four bare-bones algorithms.

In general, the aforementioned experimental results show that GBDE and other bare-bones schemes have comparable performance on all the benchmark functions, which preliminarily confirms the existence of an effective basic framework of an optimization system in solving those optimization problems.

\begin{figure}[!h]
\centerline{\includegraphics[width=8cm, angle=0]{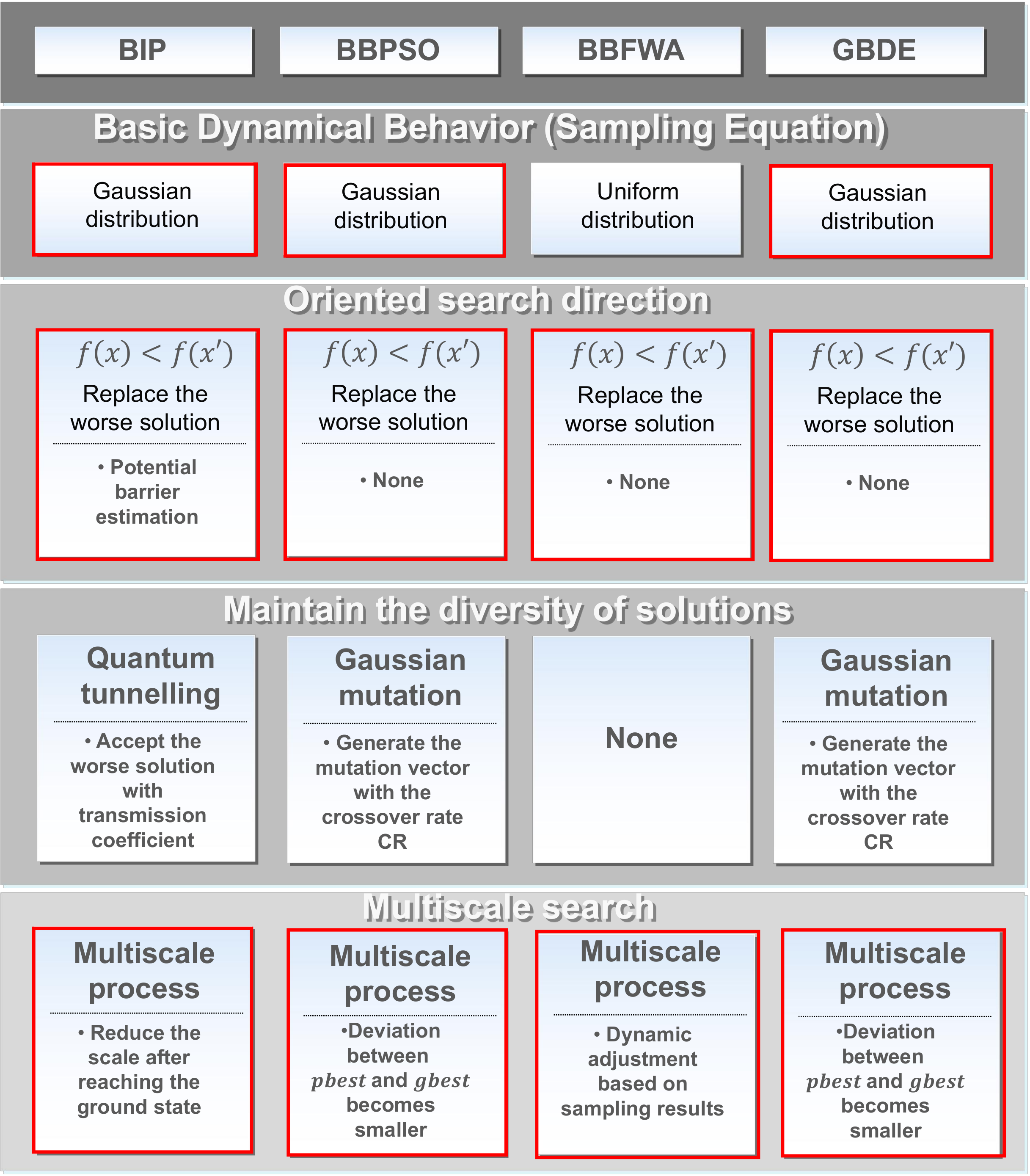} }
\caption{Schematic diagram of search mechanism between four bare-bone algorithms }
\label{fig:Compare}
\end{figure}

Based on the experimental results, we discuss the aforementioned methods in detail in the following aspects shown in Fig.~\ref{fig:Compare}.

\textit{Basic dynamic behaviour}: Among the four basic algorithms, there are three sampled Gaussian samplings shown as the basic dynamic behaviour of the algorithm. Although the constraints that affect the performance of the algorithm vary, the sampling method of the algorithm is obviously a major factor that determines the operating efficiency of the algorithm. In general, according to the experimental results, BIP, BBPSO and GBDE with Gaussian distributions have higher calculation success rates (multimodal functions) and higher calculation accuracies (unimodal functions) than BBFWA with uniform distributions.

In Tan's work, uniform distribution was inherited from the dynamic search fireworks algorithm (dynFWA) \cite{zheng2014dynamic} to enhance the randomness of the algorithm. In Kennedy's work, the Gaussian distribution centered midway between the two previous
best points and extending symmetrically beyond them, which is the famous "overflying" effect that is known to be important in exploring new regions \cite{kennedy2003bare}. In this work, the Gaussian distribution is derived from Green's function, which is the basic motion behaviour without preparatory conditions for the optimization problem. Although the researchers of the above algorithm have given their own explanations for the basic sampling behaviour, it is obvious that the explanation given in this work based on the optimization problem is more reasonable.

\textit{Oriented search direction}: In all bare-bone algorithms, replacement is performed when the newly generated solution is better than the old solution. It is interesting that few researchers have explained this behaviour of these schemes, and it seems reasonable to move to a position with a low fitness value. In this work, the movement of particles is the behaviour under the influence of the potential energy field. When the first-order Taylor expansion is employed to estimate of the objective function, the direction and probability of the movement of the particle are determined.

\textit{Diversity of solutions}: Usually, most algorithms have some strategies to maintain the diversity of solutions. In BBPSO and GBDE, Gaussian mutation is employed to maintain the diversity of solutions by probabilistically accepting the dimensional information of the mutation vector. This type of operation does not exist in BBFWA. In this work, based on the second-order Taylor expansion, the worse solution is probabilistically accepted. However, based on the experimental results, the frequency of receiving the difference solution is still high in BIP, which leads to the low intensity of the algorithm's exploration of the dominant area. In contrast, in BBPSO and GBDE, the global optimal solution has a guiding effect on all sampling points, regardless of whether it is probabilistic reception. This also leads to the performance of BBPSO and GBDE being significantly better than BIP on the multimodal function. When the number of samples of each particle in BIP is increased, the disadvantages caused by reception of worse solutions are effectively suppressed, and the performance on multimodal functions is also effectively improved (because of the limited space of this paper, the experimental data of the improved algorithm are not provided).

\textit{Multi-scale search process}: Multi-scale strategies are used in all four bare-bones algorithms. In the BBFWA, the search scale is dynamically adjusted according to the quality of the sampled results. In BBPSO and GBDE, the search scale is gradually reduced in the optimization process, which is relevant to the deviation between $gbest$ and $pbest$. In BIP, the search scale is reduced after the ground-state criteria are reached. The scales of BBFWA, BBPSO and GBDE are adaptive, but the performance of BBFWA for multimodal functions is still weaker than that of BBPSO and GBDE. This is because BBFWA lacks information sharing between particles rather than a defect of the scaling strategy. In BIP, because the ground state convergence condition is not able to adapt to all the landscape of the objective function, the iterations at each scale are not necessarily sufficient, so the algorithm performs poorly in multimodal functions and has difficulty finding the interesting area.

In summary, the four bare-bones algorithms have some similarities. Compared with other optimization models, BIP provides a more complete physical model, which supports the performance analysis and improvement of the algorithm.

\section{Conclusion}
\label{sum}

In this paper, by establishing the dynamic equation of the optimization system, the optimization problem is transformed into a constrained-state quantum problem with the objective function as the potential energy. The Schr\"odinger equation is employed to analysis the search operation under the different levels of approximation. The basic dynamic search mechanism of the intelligent optimization from quantum perspective is provided, which describes some similar search operation in comparison with other bare-bones schemes. This work demonstrates that the quantum theroy can be employed to study intelligent optimization and provide a quantum interpretation of the basic mechanism of intelligent optimization from a new perspective.

\section*{Acknowledgment}

This work was supported by the National Natural Science Foundation of China under Grant No. 60702075, Fundamental Research Funds for the Central Universities of China under Grant No. 2018NQN55, and Project of Sichuan Education Department under Grant No. 18ZB0623.

\clearpage

\bibliographystyle{elsarticle-num}
\bibliography{ref}


\end{document}